\pdfoutput=1

\documentclass{article} 

\usepackage[final]{colm2026_conference}
\usepackage{microtype}

\usepackage{algpseudocode}
\usepackage{algorithm}
\usepackage{amsfonts}
\usepackage{amssymb}   
\usepackage{latexsym}

\usepackage[most]{tcolorbox}
\usepackage{listings}
\usepackage{xcolor}

\newif\ifshowablationtable
\showablationtablefalse

\definecolor{sysAccent}{HTML}{5F5E5A}
\definecolor{neuroAccent}{HTML}{1F4E79}
\definecolor{cogAccent}{HTML}{C0641E}
\definecolor{toolAccent}{HTML}{2E8B22}

\lstdefinestyle{promptstyle}{
  basicstyle=\ttfamily\footnotesize,
  breaklines=true,
  breakatwhitespace=false,
  columns=fullflexible,
  keepspaces=true,
  showstringspaces=false,
  breakindent=0pt,
  postbreak=\mbox{\textcolor{gray}{$\hookrightarrow$}\space},
  upquote=true,
  inputencoding=utf8,
  extendedchars=true,
  literate=
    {—}{{---}}1
    {–}{{--}}1
    {→}{{$\rightarrow$}}1
    {⚠}{{!}}1
    {√}{{$\surd$}}1
    {π}{{$\pi$}}1,
}

\newcommand{\promptboxsys}[3]{%
  \begin{tcolorbox}[enhanced, breakable,
    colback=#2!4, colframe=#2!85!black,
    coltitle=white, fonttitle=\bfseries\small,
    boxrule=0.6pt, arc=2pt,
    left=6pt, right=6pt, top=4pt, bottom=4pt,
    title={#1}]
  \lstinputlisting[style=promptstyle]{#3}
  \end{tcolorbox}%
}

\newcommand{\promptbox}[3]{%
  \begin{tcolorbox}[enhanced, 
    colback=#2!4, colframe=#2!85!black,
    coltitle=white, fonttitle=\bfseries\small,
    boxrule=0.6pt, arc=2pt,
    left=6pt, right=6pt, top=4pt, bottom=4pt,
    title={#1}]
  \lstinputlisting[style=promptstyle]{#3}
  \end{tcolorbox}%
}

\usepackage[T1]{fontenc}

\usepackage[utf8]{inputenc}
\usepackage{amsmath}
\usepackage{microtype}

\usepackage{inconsolata}
\usepackage{pifont}
\usepackage{booktabs}
\usepackage{bm}
\usepackage{pifont}
\usepackage{xcolor}

\newcommand{\cmark}{\textcolor[HTML]{1B9E3F}{\ding{51}}}  
\newcommand{\xmark}{\textcolor[HTML]{D62728}{\ding{55}}}  
\newcommand{\pmark}{\textcolor[HTML]{E8820C}{\ding{108}}} 
\usepackage{tikz}
\usetikzlibrary{arrows.meta, positioning, calc}
\usepackage{graphicx}
\usepackage{xcolor}

\usepackage[table]{xcolor}          

\definecolor{oursrow}{HTML}{EDE9FE} 
\definecolor{ourtext}{HTML}{5B21B6} 
\usepackage{booktabs, multirow, bm}
\usepackage{booktabs,colortbl} 
\usepackage[dvipsnames]{xcolor}
\usepackage{rotating}    
\usepackage{booktabs}    
\usepackage{pdflscape}   
\definecolor{myred}{RGB}{220,38,38}
\definecolor{mygreen}{RGB}{22,163,74}
\definecolor{ourtext}{RGB}{80,0,200}

\usepackage{url}
\usepackage{lineno}          
\usepackage{hyperref}        
\definecolor{darkblue}{rgb}{0, 0, 0.5}
\hypersetup{colorlinks=true, citecolor=darkblue, linkcolor=darkblue, urlcolor=darkblue}

\title{NeuReasoner: Theory-grounded Mapping of Reasoning Elicitation Boundaries}

\author{%
Aydin Javadov\textsuperscript{$*$1}, Shyngys Aitkazinov\textsuperscript{$*$1}, Tobias Hoesli\textsuperscript{1},\\
\textbf{Florian von Wangenheim\textsuperscript{1}, Bj\"orn Schuller\textsuperscript{2,3}, Joseph Ollier\textsuperscript{1}}\\[0.6em]
\textsuperscript{1}ETH Z\"urich\quad
\textsuperscript{2}Imperial College London\quad
\textsuperscript{3}Technical University of Munich\\[0.3em]
}
\begin{document}

\ifcolmsubmission
\linenumbers
\fi

\renewcommand{\arraystretch}{0.96}

\maketitle


\begin{abstract}

A growing body of work suggests that the reasoning capabilities of large
language models are largely latent in their base form, with post-training
primarily amplifying rather than introducing them. However, this evidence
comes mainly from mathematical and coding benchmarks, leaving the boundary conditions of that claim---which cognitive tasks can be recovered through elicitation, and where that recovery fails---largely unexplored. To investigate this, we introduce NeuReasoner, a theory-grounded elicitation instrument. At each step, an orchestrator pairs a Neuro Lens, inspired by functional specificity, with a Cognitive Lens, drawn from the Erotetic Theory of Reasoning, and integrates their outputs through internal modularization of a single model, without external tools. We evaluate NeuReasoner on CogBench, a suite of behavioral tasks from cognitive psychology, alongside standard mathematical and coding benchmarks, measuring both its improvement over vanilla inference and its
ability to match a model's post-trained ``thinking mode.'' At sufficient
scale, NeuReasoner matches or exceeds thinking-mode baselines on arithmetic
reasoning, code generation, Bayesian reasoning, and reward learning; these
gains persist against self-consistency and iterative-refinement baselines
matched to NeuReasoner's per-decision call budget. Using NeuReasoner allows us to find clear boundaries: risk-taking and decision making under uncertainty remains hard to recover through elicitation alone, and model scale interacts with elicitation in both directions: widening its advantage on some cognitive signatures while erasing it on others. Overall, through NeuReasoner as a modular, interpretable, theory-grounded elicitation instrument, we empirically map where reasoning elicitation succeeds and fails, beyond the mathematical and coding benchmarks where prior claims have rested.
\end{abstract}

\begin{figure*}[t]
    \centering
    \includegraphics[width=0.75\textwidth]{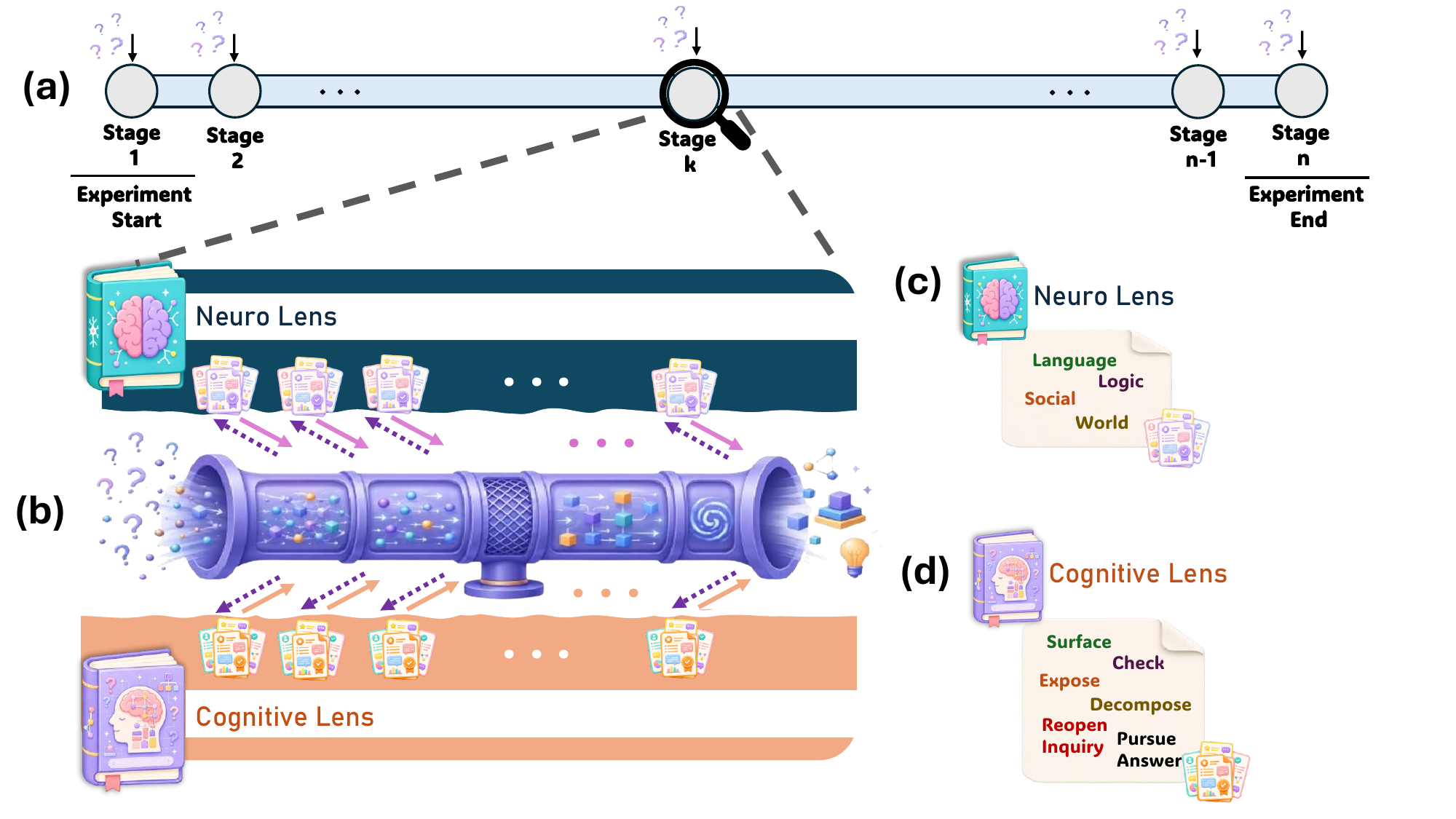}
    \caption{Overview of the NeuReasoner.
    \textbf{(a)} A CogBench \citep{codaforno2024cogbenchlargelanguagemodel} experiment runs as a sequence of stages, each
    presenting one question to solve; we refer to each such stage as a
    \emph{node}. \textbf{(b)} Within a single node (\emph{Stage k}), the
    LLM acts as an \emph{orchestrator} and, at each step, pairs one
    \emph{Neuro Lens} with one \emph{Cognitive Lens}. The two lenses are
    executed in isolation against the original question, and their
    structured outputs are integrated back into the orchestrator before
    the next step; each lens is a specialized prompt rubric drawn from a
    fixed catalog. The node ends when the orchestrator commits a final
    answer or the step budget is reached. \textbf{(c)} The Neuro Lens
    catalog: four modes of attention operationalized from large-scale
    brain networks (Language, Logic/multiple-demand, Social/theory-of-mind,
    World/default-mode). \textbf{(d)} The Cognitive Lens catalog: six
    ETR-grounded reasoning moves (Surface, Expose, Decompose, Pursue
    Answer, Check, Reopen Inquiry). For illustrative purposes, \autoref{fig:trace-example} shows a trace of a node on a sample Probabilistic Reasoning problem.}
    \label{fig:framework}
\end{figure*}

\section{Introduction}

The pursuit of robust reasoning in large language models (LLMs) has unfolded through several overlapping phases. First, the foundational work of
\citet{wei2023chainofthoughtpromptingelicitsreasoning} established that intermediate reasoning steps, known as Chains-of-Thought (CoT), can be elicited through prompting alone, opening a line of research into more elaborate reasoning schemes such as Tree-of-Thoughts \citep{yao2023treethoughtsdeliberateproblem} and Graph-of-Thoughts \citep{Besta_2024}; a line since organized into a taxonomy of reasoning topologies. Second, and in parallel, reinforcement learning (RL), a technique which has already proven
essential for aligning LLMs with human instructions \citep{ouyang2022traininglanguagemodelsfollow},  was shown to substantially improve CoT-based reasoning by rewarding correct intermediate reasoning trajectories and final answers. Subsequently, the introduction of GRPO by
\citet{shao2024deepseekmathpushinglimitsmathematical} and its adoption by \citet{Guo_2025} then consolidated RL's central role in the post-training pipeline for CoT-based reasoning.

A more recent emerging third line of work, however, has challenged existing assumptions in extant research: asking whether this progress reflects
genuinely new capability or simply the amplification of competences that already exist in the base model. A growing body of findings converges toward the latter. \citet{liu2025understandingr1zeroliketrainingcritical}, for example, observed that widely used base models already display strong reasoning behavior spontaneously, including the ``aha moment'' self-reflection patterns taken as signatures of emergent reasoning. Likewise, \citet{yue2025doesreinforcementlearningreally} showed that the reasoning traces produced by RL-fine-tuned models were already present among a base model's own generations once it is sampled sufficiently. A related
framing of the same observation, \citet{he2025rewardingunlikelyliftinggrpo,shao2026spuriousrewardsrethinkingtraining, yue2025doesreinforcementlearningreally} describes this as \emph{distribution sharpening}, where the post-trained distribution is viewed as a sharper version of the base model's own distribution, concentrating mass on traces it could already produce rather than on ones it would otherwise be unlikely to generate. Consistent with this picture, \citet{ebouky2025elicitingreasoninglanguagemodels} show that equipping a base LLM with self-executed ``cognitive tools''
yields substantial gains on standard mathematical benchmarks, and \citet{karan2025reasoningsamplingbasemodel} show that additional inference-time sampling alone can lift a base model to the single-shot performance of its RL post-trained counterpart. The collective message from this literature, therefore, is that the base model is smarter than it appears; the question we take up in this paper is \emph{how much} smarter, and on \emph{what kinds} of reasoning tasks.

Despite making important contributions, in addressing this question, two clear limitations exist in literature to date. First,
this claim rests on a narrow empirical foundation, which are tests performed almost entirely on
mathematical and coding benchmarks. While this is in part a natural consequence of
how the underlying RL methods work: equipped with reward signals that
are cheap to verify automatically, with RL post-training widely applied
 successfully in domains like mathematics, coding, and science \citep{hendrycks2021measuringmathematicalproblemsolving, rein2023gpqagraduatelevelgoogleproofqa, Li_2022}, this means the elicitation work that responds to it has inherited similar benchmarks. Consequently, this limits generalizability of claims regarding distribution sharpening. 

Second, while this work borrows the vocabulary of cognition, the grounding is largely atheoretical: its building blocks are task-decomposition heuristics
with cognitively flavored names rather than commitments to any specific theory of how humans reason. This has led to an intuitive, but in some respects shallow, approach to explaining model reasoning, which could benefit from a more structured, theoretically grounded approach, robust across the diverse types of reasoning tasks with which humans regularly engage.


To close the gap between cognitive vocabulary used to describe current model reasoning and the actual cognitive structure of human thinking, we introduce a framework that integrates theories from cognitive science and neuroscience directly into the agent's step-by-step reasoning. Our goal is to test of whether theory-constrained modular elicitation can recover task-specific latent reasoning behavior, and where that recovery fails.

\paragraph{Our contributions.}
\begin{itemize}

\item Providing, to our knowledge, one of the first behavioral and psychologically grounded characterizations of where reasoning elicitation from base models succeeds and where it breaks down. We show that while elicitation can recover strong reasoning performance in several settings, its ability to match post-trained ``thinking'' models is task-contingent. In particular, tasks that centrally involve decision-making under uncertainty, such as the Restless Bandit Experiment and the Balloon Analog Risk Task, expose a consistent gap: thinking models outperform NeuReasoner. This identifies a concrete boundary condition for reasoning elicitation beyond standard mathematical and coding benchmarks.

\item NeuReasoner, the interpretable elicitation instrument developed to establish the above findings: a neuro-cognitively grounded approach that combines erotetic reasoning theory with functional-specificity literature to align an LLM's step-by-step problem solving with cognitive-scientific decompositions of human cognition, using only internal modularization and being training-free.

\end{itemize}

\section{Background \& Related Work}


In this section, we outline the established cognitive and neuroscience foundations that motivate both our reasoning framework and the behavioral evaluations used in its assessment. This includes use of the Erotetic Theory of Reasoning (ETR) to explain how humans approach tasks as question-drive enquiry; how functional specificity from neuroscience explains specialized modes of reasoning; and how findings on cognitive architectures motivate decomposing reasoning into modular operations.

\paragraph*{The Erotetic Theory of Reasoning.} A recurring theme in modern accounts of human reasoning is its duality: competence and predictable error go hand in hand \citep{richardson2026theorygroundedevaluationhumanlikefallacy}.
Across many domains, people deviate from normative reasoning along a stable, repeatable set of fallacies \citep{doi:10.1126/science.185.4157.1124, kahnemann1990, evans1989bias, JOHNSONLAIRD200627}. Consequently, a cognitively grounded system must anticipate weaknesses of human reasoning as deliberately as strengths. The Erotetic Theory of Reasoning (ETR) \citep{koralus2013} accounts for this, denoting that reasoning is not the manipulation of propositions but rather an inquiry-driven process: a reasoner maintains a set of disjunctive alternatives and filters them as new information arrives, in order to resolve an implicit question. This question-driven process is what makes the theory useful for our purposes. On the one hand, ETR usually surfaces the most probable answer efficiently; on the other, the same mechanism can narrow the space of alternatives prematurely and produce predictable errors 
\citep{richardson2026theorygroundedevaluationhumanlikefallacy}. Both behaviors are valuable: the first is the competence our framework aims to elicit, while the second yields interpretable failure modes.

\paragraph*{Functional Specificity.}
We follow neuro- and cognitive-science evidence that human reasoning is supported not by a homogeneous mechanism, but by specialized networks preferentially recruited for different cognitive demands \citep{Kanwisher2010}. While \citet{alkhamissi_mixture_2025} use this insight in a mixture-of-experts architecture, we ask how functional specificity can inform \emph{elicitation}. Following their decomposition, our design mirrors four canonical networks: the language network \citep{fedo2011}, multiple-demand network \citep{duncan2010}, theory-of-mind network \citep{SAXE20031835}, and default-mode network \citep{gusnard}.

\paragraph*{Cognitive Architectures and Modular Reasoning.}
Another line in cognitive science treats reasoning not as a
single faculty but as the orchestrated interplay of distinct mental operations. Cognitive architectures such as ACT-R \citep{Anderson01121997} make this explicit:
modeling higher-level thought and its links to perception as the
emergent behavior of separable components for goal management,
procedural skill, and memory retrieval. The same decompositional
stance has recently been carried into work on LLMs.
\citet{ebouky2025elicitingreasoninglanguagemodels} cast individual
reasoning operations as callable cognitive tools, while
\citet{sumers2024cognitivearchitectureslanguageagents} placed the LLM
as a central controller over modular memory and a structured action
space, separating internal actions such as reasoning and retrieval
from external ones such as API calls. What unites these efforts is the
premise that reasoning is better staged than collapsed into one
monolithic pass. This also brings a concrete benefit, as in biological and artificial systems alike, modular
organization has been tied to compositional generalization, the
ability to recombine familiar operations to solve unfamiliar problems
\citep{ito2022compositionalgeneralizationabstractrepresentations}. This decompositional position has independent empirical support beyond text: in multimodal reasoning, \citep{yucheng} identify a task-composition bottleneck, where recognition and reasoning cannot be carried out jointly in a single pass, and show that explicitly decoupling the two stages recovers performance; showing direct evidence that staging, rather than collapsing the reasoning operations matters.


\begin{table*}[t]
\centering
\small
\setlength{\tabcolsep}{2pt}
\renewcommand{\arraystretch}{1.3}
\begin{tabular}{l c c c c c c c}
\toprule
\textbf{Work} & \textbf{Train-Free} & \textbf{Beyond} & \textbf{Self-} &
\textbf{Modular} & \textbf{Neuro} & \textbf{CogSci} & \textbf{Elicitation} \\
 & \textbf{Elicitation} & \textbf{Math \& Code} & \textbf{Agency} &
 & \textbf{Inspired} & \textbf{Inspired} & \textbf{Scope} \\
\midrule
\citet{ebouky2025elicitingreasoninglanguagemodels}      & \cmark & \xmark & \cmark & \cmark & \xmark & \cmark & \pmark \\
\citet{alkhamissi_mixture_2025}           & \xmark & \cmark & \xmark & \cmark & \cmark & \cmark & \xmark \\
\citet{karan2025reasoningsamplingbasemodel} & \cmark & \pmark & \xmark & \xmark & \xmark & \xmark & \xmark \\
\citet{codaforno2024cogbenchlargelanguagemodel}     & \cmark & \cmark & \xmark & \xmark & \xmark & \cmark & \pmark \\
\citet{kramer2024unlockingstructuredthinkinglanguage} & \cmark & \xmark & \pmark & \cmark & \xmark & \cmark & \xmark \\
\citet{sumers2024cognitivearchitectureslanguageagents}              & \xmark & \cmark & \xmark & \cmark & \xmark & \cmark & \pmark \\
\midrule
\textcolor{ourtext}{\textbf{NeuReasoner}}                   & \cmark & \cmark & \cmark & \cmark & \cmark & \cmark & \cmark \\
\bottomrule
\end{tabular}
\caption{Positioning of \textit{\textbf{NeuReasoner}} relative to prior work.
\cmark{}~=~yes, \pmark{}~=~partial, \xmark{}~=~no. \emph{Train-Free
Elicitation}: improves reasoning with no additional training.
\emph{Beyond Math \& Code}: evaluated outside mathematical and coding
benchmarks. \emph{Self-Agency}: the model orchestrates its own
reasoning internally, with operations executed by the same model and
no external tool-calling or environment-interaction harness.
\emph{Modular}: reasoning is decomposed into distinct operators or
components. \emph{Neuro / CogSci Inspired}: grounded in neuroscientific
/ cognitive-science theory. \emph{Elicitation Scope}: studies
\emph{where} and \emph{how far} elicitation works, not only
\emph{whether} it does.}
\label{tab:related}
\end{table*}


A growing body of work has sought to improve LLM reasoning without
additional training, with a subset drawing from
cognitive and neuroscience. To position the uniqueness of our framework against existing
approaches, we classify prior work along seven dimensions in Table~\ref{tab:related}.
Our framework is, to our knowledge, the first to combine all
seven: a train-free, internally orchestrated, theory-grounded
elicitation method evaluated beyond mathematical reasoning, with the
limits of elicitation treated as a primary research question rather
than a by-product.

\section{Method}
\label{sec:method}

\paragraph{Framework Design}
\label{sec:framework-design}

Each CogBench \citep{codaforno2024cogbenchlargelanguagemodel}
experiment unfolds as a sequence of stages, and at every stage the
model is presented with a question to solve, with the question's form
depending on the experiment type (Figure~\ref{fig:framework}). We refer
to each such experiment stage as a \emph{node}, the unit on which our
framework operates, and to each orchestrator decision taken within a
node as a \emph{step}. Our elicitation framework operationalizes
reasoning at the level of the individual node, where the
micro-decisions of solving a single question are made. Without the loss of generality, for the math and code generation evaluations, $n=1$. 

\paragraph{Formalization.}
Let $q$ denote the question at a node and $\pi$ the single language
model used throughout, queried in different roles. We write the two
operator catalogs as $\mathcal{N}$, the set of Neuro Lenses, and
$\mathcal{C}$, the set of Cognitive Lenses. A node is solved by
iterating a state $h_t$, the reasoning history available to the
orchestrator at step $t$, with $h_0$ initialized from $q$ and the
system rules. At each step, the orchestrator produces a decision
\begin{equation}
s_t \;=\; \pi_{\text{orch}}(h_t)
\;\in\;
\big(\mathcal{N}\times\mathcal{C}\big)
\;\cup\;
\{\textsc{commit}(a)\},
\end{equation}
that is, $s_t$ is either an \emph{intermediate} step, which selects a
Neuro Lens $n_t$ and a Cognitive Lens $c_t$, or a \emph{terminal} step
$\textsc{commit}(a)$ that emits a final answer $a$ and ends the node.
Together with the lens pair the orchestrator also issues a
\emph{directive} $d_t$, a short natural-language instruction that
tells the two chosen lenses what to attend to at this step: the lens
prompt fixes the expert \emph{role}, while $d_t$ fixes the
step-specific \emph{task} given to that role. For an intermediate
step, the two lenses are executed independently as further calls to
the same model,
\begin{equation}
o^{\mathcal{N}}_t = \pi_{n_t}(q, d_t, r_t),
\qquad
o^{\mathcal{C}}_t = \pi_{c_t}(q, d_t, r_t),
\end{equation}
where $r_t$ is an optional reference to a previous step, and neither
call observes the other's output. The orchestrator then integrates the
two structured outputs into the next state,
$h_{t+1} = h_t \cup \{o^{\mathcal{N}}_t, o^{\mathcal{C}}_t\}$, and the
loop continues. The node terminates at the first terminal step, or is
forced to terminate once a step budget $T$ is reached, so that every
node halts within at most $T$ steps. Unlike cognitive prompting \citep{kramer2024unlockingstructuredthinkinglanguage}, and in line with cognitive tools \citep{ebouky2025elicitingreasoninglanguagemodels}, our framework imposes no fixed, monolithic ordering on the lenses.  In structuring reasoning as bounded calls, our procedure connects to the token- and budget-forcing view of test-time scaling proposed by \citet{muennighoff2025s1simpletesttimescaling}. 
\autoref{alg:loop} summarizes the full per-node loop.

\begin{algorithm}[t]
\caption{Per-node reasoning loop}
\label{alg:loop}
\begin{algorithmic}[1]
\Require question $q$, step budget $T$
\State $h_0 \gets$ initialize from $q$, system rules \Comment{node state}
\For{$t = 0$ \textbf{to} $T-1$}
    \State $s_t \gets \pi_{\text{orch}}(h_t)$ \Comment{pick next step}
    \If{$s_t = \textsc{commit}(a)$}
        \State \Return $a$ \Comment{answer found}
    \Else
        \State $(n_t, c_t, d_t, r_t) \gets s_t$ \Comment{unpack step}
        \State $o^{\mathcal{N}}_t \gets \pi_{n_t}(q, d_t, r_t)$ \Comment{Neuro Lens}
        \State $o^{\mathcal{C}}_t \gets \pi_{c_t}(q, d_t, r_t)$ \Comment{Cognitive Lens}
        \State $h_{t+1} \gets h_t \cup \{o^{\mathcal{N}}_t, o^{\mathcal{C}}_t\}$ \Comment{integrate}
    \EndIf
\EndFor
\State \Return $\textsc{commit}(a)$ forced from $h_T$ \Comment{budget reached}
\end{algorithmic}
\end{algorithm}


\paragraph{Cognitive Lenses.}
The six Cognitive Lenses operationalize central ideas from the erotetic theory of reasoning (ETR)
\citep{koralus2013}, where reasoning is organized around the \emph{issue} to be resolved.
They decompose reasoning into issue formulation, presupposition checking, subquestion decomposition,
candidate-answer generation, resolution checking, and inquiry revision. Full prompts are given in
Appendix~\ref{app:prompts}.

\paragraph{Neuro Lenses.}
The four Neuro Lenses operationalize broad computational roles associated with functionally
specialized brain networks. The Language Network Lens targets linguistic interpretation; the
Multiple-Demand Lens targets structured control and comparison; the Theory-of-Mind Lens targets agents' beliefs and perspectives; and the Default-Mode Lens targets internal world modeling and temporal simulation. We do not claim these lenses simulate human cognition; each only approximates the characteristic computational role its network is associated with in the literature, shaping what information a step prioritizes and which failure modes it monitors. Full prompts are given in Appendix~\ref{app:prompts}.

\begin{figure}[t]
\centering
\definecolor{neurocol}{HTML}{1F4E79}   
\definecolor{cogcol}{HTML}{C0641E}     
\definecolor{qcol}{HTML}{3A3A3A}       
\definecolor{anscol}{HTML}{1B7A3D}     
\definecolor{stepbg}{HTML}{F2F2F2}     

\resizebox{0.72\columnwidth}{!}{%
\begin{tikzpicture}[
    font=\small,
    node distance=2.0mm,
    box/.style={rounded corners=2pt, draw, align=left,
                text width=0.86\columnwidth, inner sep=4pt},
    lens/.style={rounded corners=2pt, draw=none, font=\footnotesize\bfseries,
                 text=white, inner xsep=4pt, inner ysep=1.5pt},
    flow/.style={-{Stealth[length=2mm]}, thick, gray}
]

\node[box, fill=qcol!8, draw=qcol!50] (q) {%
    \textbf{Task (Probabilistic Reasoning).}
    A wheel sends $9/10$ sections to urn~F (prior $0.9$).
    Both urns hold 5 red / 5 blue balls. A \textbf{blue} ball is drawn.
    Estimate $P(\text{urn}=\text{F})$.};

\node[box, fill=stepbg, below=of q] (s1) {%
    \tikz\node[lens, fill=neurocol]{Logic};\,$\times$\,%
    \tikz\node[lens, fill=cogcol]{Surface Issue}; \\[2pt]
    Frames the task as a Bayesian posterior. Records
    $P(F)=0.9$ and $P(\text{blue}\mid F)=P(\text{blue}\mid J)=0.5$.};
\node[anchor=north west, font=\footnotesize\bfseries,
      text=gray] at ([xshift=162pt,yshift=-1pt]s1.north west) {Step 1};

\node[box, fill=stepbg, below=of s1] (s2) {%
    \tikz\node[lens, fill=neurocol]{Logic};\,$\times$\,%
    \tikz\node[lens, fill=cogcol]{Pursue Answer}; \\[2pt]
    Applies Bayes. Since the likelihood is equal on both sides
    ($0.5/0.5$), the observation is \emph{uninformative}: the
    posterior stays at the prior.};
\node[anchor=north west, font=\footnotesize\bfseries,
      text=gray] at ([xshift=162pt,yshift=-1pt]s2.north west) {Step 2};

\node[box, fill=anscol!10, draw=anscol!60, below=of s2] (ans) {%
    \textbf{Commit:} $P(\text{urn}=\text{F})=\mathbf{0.90}$ \;
    (Bayes-optimal $=0.900$). The model recognizes the draw carries
    no information and \emph{holds the prior} rather than over-updating.};

\draw[flow] (q)  -- (s1);
\draw[flow] (s1) -- (s2);
\draw[flow] (s2) -- (ans);

\end{tikzpicture}}

\caption{A worked trace of NeuReasoner on a Probabilistic
Reasoning node (\texttt{qwen3-14B}). At each step, the orchestrator
pairs a Neuro Lens (\textcolor{neurocol}{\textbf{blue}}) with a
Cognitive Lens (\textcolor{cogcol}{\textbf{orange}}). The task
superficially invites a Bayesian update, but because the blue draw is
equally likely under both urns it is uninformative; the model
correctly holds the prior and commits the Bayes-optimal answer in two
steps.}
\label{fig:trace-example}
\end{figure}
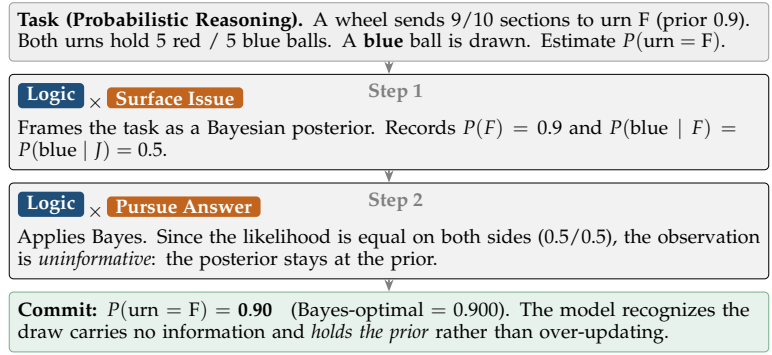

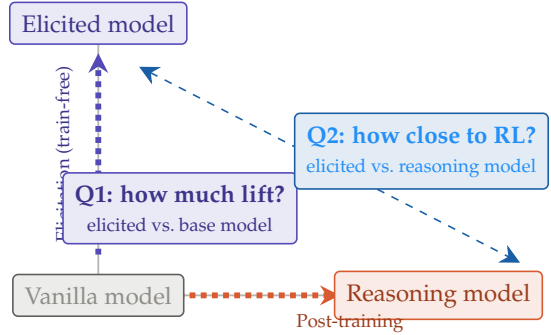
\begin{figure}[t]
\centering
\definecolor{baseCol}{HTML}{5F5E5A}
\definecolor{elicCol}{HTML}{534AB7}
\definecolor{elicTx}{HTML}{3C3489}
\definecolor{rlCol}{HTML}{D85A30}
\definecolor{rlTx}{HTML}{993C1D}
\definecolor{gapCol}{HTML}{185FA5}
\definecolor{axisCol}{HTML}{888780}

\begin{tikzpicture}[
    font=\small,
    xscale=0.9, yscale=0.72,
    >={Stealth[length=2.6mm,width=2.6mm]},
    box/.style={rounded corners=2pt, draw, line width=0.6pt,
                align=center, inner sep=4.5pt, font=\footnotesize},
]

\draw[axisCol, line width=0.7pt, opacity=0.6] (0,0) -- (0,5.3);
\draw[axisCol, line width=0.7pt, opacity=0.6] (0,0) -- (5.7,0);

\draw[elicCol, line width=2.2pt, dotted, ->] (0,0.75) -- (0,4.45);
\draw[rlCol,   line width=2.2pt, dotted, ->] (0.85,0) -- (3.55,0);

\draw[gapCol, line width=0.6pt, dashed, <->] (0.6,4.2) -- (6.2,0.6);

\node[box, fill=baseCol!12, draw=baseCol, text=baseCol!150!black]
      (base) at (0,0) {Vanilla model};
\node[box, fill=elicCol!12, draw=elicCol, text=elicTx]
      (elic) at (0,5) {Elicited model};
\node[box, fill=rlCol!12, draw=rlCol, text=rlTx]
      (rl) at (5,0) {Reasoning model};

\node[elicTx, font=\scriptsize, rotate=90] at (-0.5,2.6)
      {Elicitation (train-free)};
\node[rlTx, font=\scriptsize] at (3.7,-0.52)
      {Post-training};

\node[box, fill=elicCol!12, draw=elicCol, text=elicTx]
      at (1.2,1.6) {\textbf{Q1: how much lift?}\\[1pt]
      \scriptsize elicited vs.\ base model};
\node[box, fill=gapCol!12, draw=gapCol, text=gapCol!150!black]
      at (4.75,2.65) {\textbf{Q2: how close to RL?}\\[1pt]
      \scriptsize elicited vs.\ reasoning model};

\end{tikzpicture}

\caption{From a vanilla
model, post-training is the established route to a reasoning-capable
model; we instead study \emph{elicitation}, which seeks reasoning gains
with no parameter updates. This raises two questions, examined across a
suite of cognitive-psychology tasks: \textbf{Q1}, how much does
elicitation lift performance over the unmodified model; and
\textbf{Q2}, how close does the elicited model come to one improved by
post-training. The figure is a schematic; quantitative, per-task
results appear in \S\ref{sec:res}.}
\label{fig:two-directions}
\end{figure}



\begin{table*}[t]
\centering
\footnotesize
\setlength{\tabcolsep}{4pt}
\renewcommand{\arraystretch}{1.0}
\begin{tabular}{@{}lllr@{}}
\toprule
Experiment & Cognitive / behavioral faculty & Decision format & Runs \\
\midrule
BART                    & Risk taking                        & Option 1/2 (inflate / skip)   & 5 \\
Horizon Task            & Directed / random exploration      & Machine letter (F, J, \dots)  & 1 ($\sim$100 trials) \\
Instrumental Learning   & Learning rate; optimism bias       & Machine letter                & 8 \\
Probabilistic Reasoning & Prior \& likelihood weighting      & Probability (0.XX)            & 10 \\
Restless Bandit         & Meta-cognition                     & Machine letter                & 5 \\
Temporal Discounting    & Temporal discounting               & Option 1/2                    & 5 \\
Two-Step Task           & Model-basedness                    & Planet then alien letter      & 1 ($\sim$25 trials) \\
\bottomrule
\end{tabular}
\caption{The seven CogBench \citep{codaforno2024cogbenchlargelanguagemodel} experiments. Each produces both a task-level performance result (reward or accuracy) and one or two behavioural metrics that characterise the cognitive strategy the model uses, not just whether it succeeds. The Cognitive / Behavioral Faculty column lists those behavioural dimensions. HorizonTask and TwoStepTask are each a single extended session of $\sim$100 and $\sim$25 sequential trials respectively; all other experiments are repeated as independent runs.}
\label{tab:cogbench}
\end{table*}

%




\section{Experimental Setup}
\label{sec:setup}

\paragraph{Datasets}
\label{sec:setup-data}

\paragraph{CogBench.}
Our primary evaluation uses CogBench
\citep{codaforno2024cogbenchlargelanguagemodel}, a benchmark of seven
cognitive-psychology paradigms. Each experiment yields two classes of metric: a \emph{performance score} (task reward or accuracy) and one or more \emph{behavioral scores} extracted by fitting task-specific computational models to the full response sequence, capturing \emph{how} a model behaves rather than only whether it succeeds. Across all seven experiments this gives ten behavioral metrics directly comparable to human norms; six experiments also carry a distinct performance score (Temporal Discounting's performance metric coincides with its behavioral one). All scores are normalized so that random responding is $0$ and human performance is $1$. Table~\ref{tab:cogbench} lists the seven experiments.


\paragraph{Mathematical \& Coding benchmarks.}
As another evaluation we also use the mathematical benchmarks AIME~2024 \citep{maa_aime_problems_solutions_2024},
MATH-500 \citep{lightman2024lets}, and AMC \citep{aimo_amc_2024}, and the coding benchmark
HumanEval+ \citep{liu2023evalplus}.

We evaluate on the Qwen3 family (8B, 14B, 32B) \citep{qwen_team_2025_qwen3}, a \emph{hybrid} model whose single checkpoint runs either in a \emph{thinking} mode, which emits an extended chain-of-thought before
answering, or a \emph{non-thinking} mode that answers directly; the thinking mode is the behavior instilled during Qwen3's post-training.
This hybrid design makes Qwen3 suitable for both evaluation settings in which elicitation can be measured directly against post-training as well as the vanilla model.

\paragraph{Experiments}
\label{sec:setup-experiments}

Following \autoref{fig:two-directions}, our evaluation has two arcs.

\paragraph{The lift, on instruction-tuned models.}
For each instruction-tuned model we compared the model alone
(\emph{Vanilla}, following the original CogBench protocol) against the
neuro-cognitive framework applied to that same model
(\S\ref{sec:framework-design}). As these models have no post-trained
reasoning mode, this contrast isolates the framework's lift.


\paragraph{Lift and the thinking mode, on the Qwen3 family.}
On the Qwen3 family, we evaluated three conditions: \textbf{C1}, the
non-thinking model; \textbf{C2}, the same model with thinking mode
enabled; and \textbf{C3}, NeuReasoner applied to the non-thinking
model. The contrast C1/C3 measures NeuReasoner's lift, placing this
family in the first arc as well; C2/C3 asks whether that lift matches
the post-trained thinking mode, the question of the second arc; and
C1/C2 reports the effect of the thinking mode on its own.

\paragraph{Mathematical and coding sanity check.}
Because verifiable math and code are a regime where a code interpreter is standard, for these benchmarks, we evaluated a separate \emph{tool-augmented} variant of the framework, in which operators may also invoke a Python interpreter.\looseness=-1

\paragraph{Implementation Details}
\label{sec:setup-impl}

For all CogBench experiments we used temperature $0$, yielding
deterministic responses across every condition. For the mathematical
and coding benchmarks, we followed the Qwen3 team's recommended decoding
settings \citep{qwen_team_2025_qwen3}, temperature $0.6$ in thinking mode and $0.7$ in non-thinking mode. Detailed cost breakdowns and prompts are in Appendix \ref{app:cost} and \ref{app:prompts}.

\section{Results and Discussion}
\label{sec:res}
\paragraph{NeuReasoner matches the thinking mode on math and code at scale}
\label{sec:res-math}

Evaluation on AIME, AMC, MATH-500, and HumanEval+ (all Pass@1) comparing \emph{Qwen3-32B (thinking off)}, \emph{Qwen3-32B (thinking on)}, and NeuReasoner reveals that NeuReasoner matches or exceeds the thinking mode on AIME, AMC, and MATH-500, trailing only narrowly on HumanEval+ (87.6\,\% vs.\ 88.9\,\%, within SEM). The full per-model table is given in Appendix~\ref{app:math-code}.

\paragraph{Reasoning Elicitation beyond Arithmetic. (When) Does Elicitation via NeuReasoner Lift Reasoning?}
\label{sec:res-lift}

\begin{figure*}[htbp]
    \centering
    \includegraphics[width=0.75\linewidth]{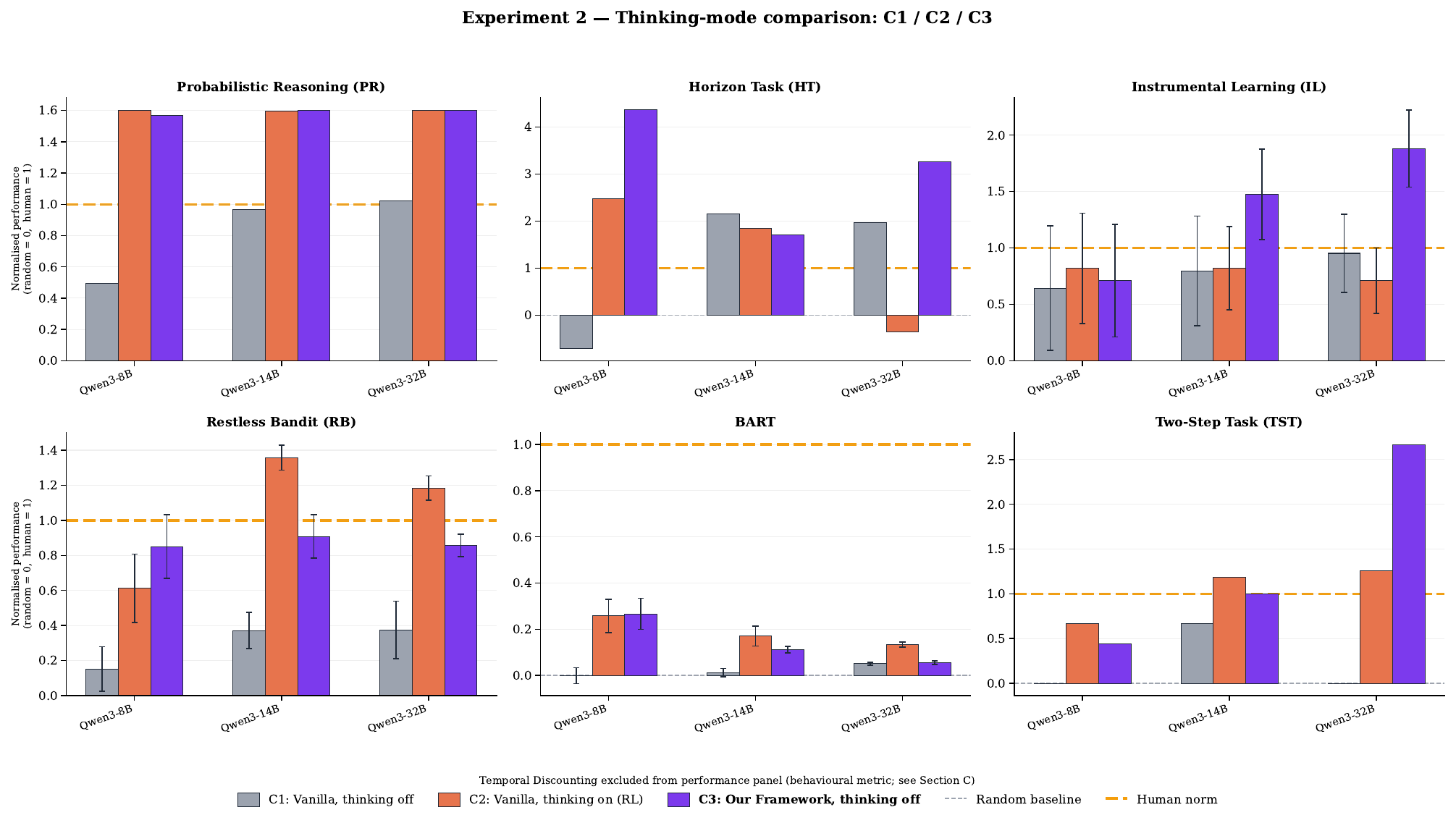}
    \caption{Normalized performance (random\,=\,0, human\,=\,1) across six CogBench performance tasks comparing three conditions on Qwen3-\{8B,\,14B,\,32B\}: \textbf{C1}~vanilla with thinking disabled, \textbf{C2}~vanilla with post-trained chain-of-thought enabled, and \textbf{C3}~NeuReasoner (structured elicitation, thinking disabled). Temporal Discounting is excluded from this panel as its performance metric coincides with its behavioral score (see \autoref{tab:phenotype_compact}). The amber dashed line marks human-level performance. Error bars denote $\pm$1\,SEM; hatching indicates partial runs. C3 matches or exceeds C2 on the majority of tasks without any reinforcement-learning fine-tuning, most consistently on Bayesian inference and reward learning. Per-decision cost is analyzed in \autoref{app:cost}.}
    \label{fig:exp2_performance}
\end{figure*}

\autoref{fig:exp2_performance} shows results across CogBench: NeuReasoner beats the vanilla baseline in nearly all experiments across model sizes, with Qwen3-14B as an exception for the Temporal Discounting and Directed Exploration tasks, which are core computational concepts in behavioral science and reinforcement learning that explain how humans value rewards across time and navigate uncertainty.

\paragraph{(When) Does Reasoning Elicitation via NeuReasoner Reach What Post-Training Reaches?}

\begin{table}[t]
  \centering
  \caption{Normalized performance averaged across all 7 CogBench experiments. Values are mean\,$\pm$\,SEM.}
  \label{tab:summary_conditions}
  \resizebox{\columnwidth}{!}{%
  \begin{tabular}{l r r r r}
    \toprule
    \textbf{Condition} & \textbf{Qwen3-8B} & \textbf{Qwen3-14B} & \textbf{Qwen3-32B} & \textbf{Avg} \\
    \midrule
    Baseline
      & $-0.410 \pm 0.531$ & $+1.176 \pm 0.430$ & $+0.013 \pm 0.760$
      & $+0.260$ \\
    Thinking ON (RL)
      & $+1.027 \pm 0.287$ & $\mathbf{+1.343 \pm 0.274}$ & $+0.395 \pm 0.443$
      & $+0.921$ \\
    \rowcolor{oursrow}
    \textcolor{ourtext}{\textbf{NeuReasoner}}
      & $\mathbf{+1.519 \pm 0.552}$ & $+1.319 \pm 0.276$ & $\mathbf{+1.821 \pm 0.416}$
      & $\mathbf{+1.553}$ \\
    \bottomrule
  \end{tabular}
  }
\end{table}

Overall, NeuReasoner applied to vanilla, is on par or better than the ``thinking'' model, having the highest overall performance among the three conditions, across the model family (see ~\autoref{tab:summary_conditions}), and specifically (see ~\autoref{fig:exp2_performance}), for probabilistic reasoning, directed exploration, temporal discounting, and reward learning. It struggled, however, on Restless Bandit and the Balloon Analog Risk Task, evaluating uncertainty, monitoring outcomes, and regulating behavior.
This point is visually displayed in radar plots of phenotype in Appendix, ~\autoref{fig:exp3_phenotype_radar}, which compare the
three conditions across Qwen3-\{8B,\,14B,\,32B\}. NeuReasoner and thinking model reshape the behavioral fingerprint similarly on deliberation dimensions, but diverge on exploration and heuristic dimensions, suggesting that structured elicitation and post-training affect different cognitive facets.

\begin{figure}[t]
  \centering
  \includegraphics[width=0.7\columnwidth]{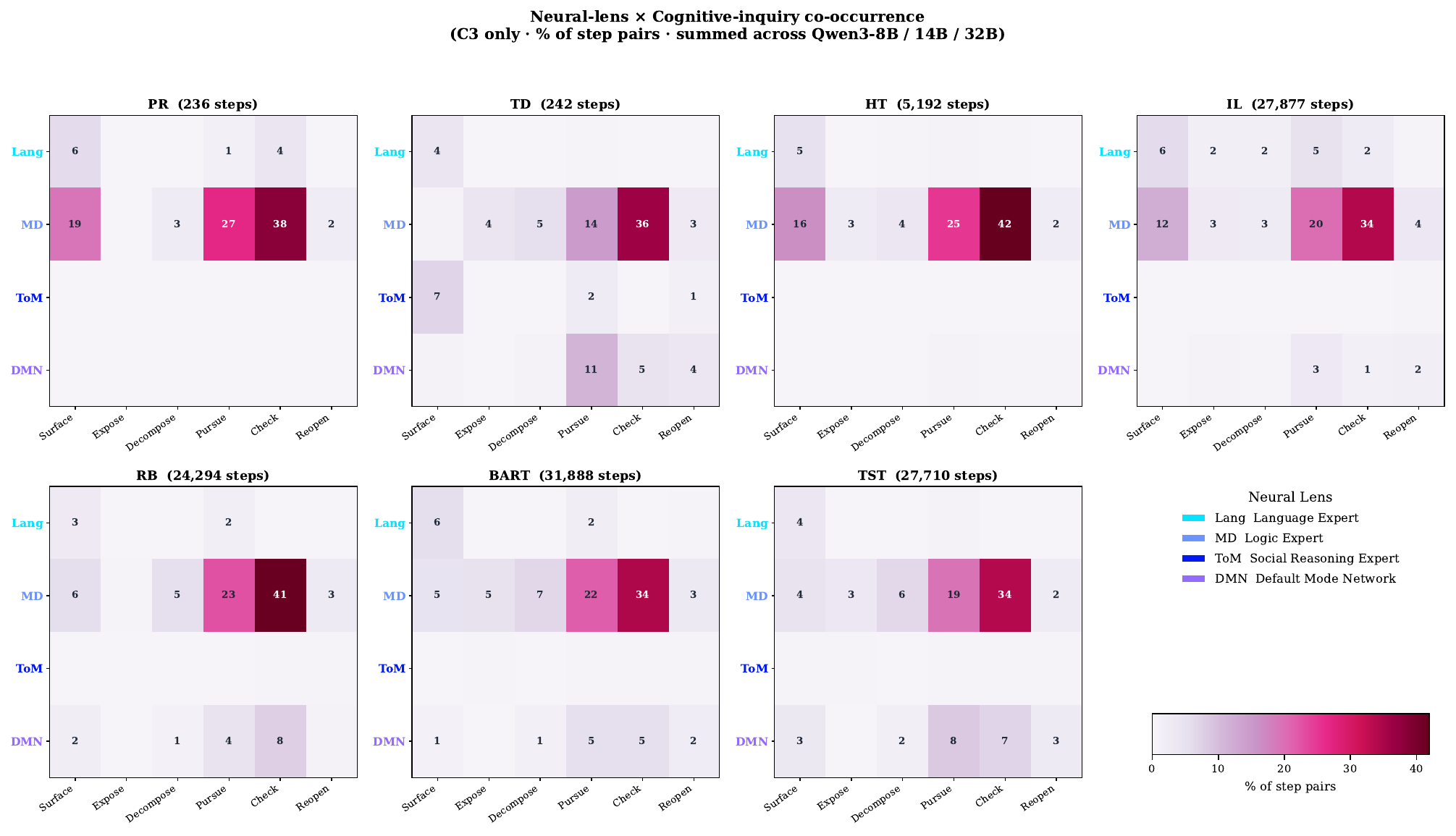}
  \caption{\textbf{Brain-lens $\times$ cognitive-inquiry co-occurrence, summed across models.}
  Each heatmap shows the percentage of reasoning steps that used each (Neuro Lens, Cognitive Lens) pair,
  broken down by model and CogBench experiment.
  Cell values $\geq$\,1\% are annotated.
  The concentration of mass in one or two cells per model confirms that   models do not distribute their use across the full theoretical catalog.}
  \label{fig:lens_heatmap_exp1-body-2}
\end{figure}

\paragraph{Are the gains structure or compute?}
\label{sec:res-compute}

NeuReasoner issues 11--39 LLM calls per decision (\autoref{app:cost}), raising
the question of whether its gains reflect the lens structure or merely the
extra inference-time compute. To separate these, we ran two compute-matched
baselines on Qwen3-8B and 32B at NeuReasoner's per-task call budget:
self-consistency (SC; $N$ samples, majority vote) and iterative refinement
(IR; $N$ sequential re-examinations). Full tables are in
\autoref{app:compute-matched}.

Matched compute does not substitute for structure. On 8B, SC and IR recover
only part of NeuReasoner's advantage on five of seven tasks and fail to
reproduce the diagnostic cognitive signatures, e.g., both show zero directed
exploration on the Horizon Task, and SC falls below random on Two-Step
model-basedness even while matching reward, as majority voting erases the
sequential dependency model-based planning requires. The two partial
exceptions at 8B (Two-Step reward, Instrumental Learning) both reverse at 32B,
where SC and IR drop to or below random while NeuReasoner holds or widens its
lead. Scale cuts both ways, however: the 8B directed-exploration signature
vanishes at 32B, where a strong exploitation prior leaves all conditions below
random, bounding \emph{when} elicitation can reshape a signature.

\paragraph{Lens pick flexibility.}

NeuReasoner's flexibility in selecting which neuro- and cognitive lenses to apply to a given problem also enables post-hoc interpretability and auditability of the resulting reasoning traces. 
We next ask: to what extent does this flexibility manifest in the statistics of lens-pair selections across tasks and models?

 \autoref{fig:lens_heatmap_exp1-body-2} shows how often each (Neuro Lens, Cognitive Lens) pair is selected together, and \autoref{fig:lens_heatmap_exp1-body} shows a summary of the Brain × Cognitive lens selection
averaged across models for each CogBench experiment. The full numerical breakdown per model and experiment is given in ~\autoref{tab:cog_neural_per_task_model}.

The concentration of mass confirms that models do not uniformly distribute their use across the full theoretical catalog, which raises questions on the mechanistic insights of model behavior, out of scope of the current paper. Yet, our ablation studies confirm that, overall, all lenses made a contribution towards the reasoning elicitation behavior of NeuReasoner (\autoref{fig:ablation_summary}). We report the full sweep in ~\autoref{app:ablations}.




\section{Conclusion}

We set out to map where reasoning elicitation from base models succeeds and where it breaks down, developing NeuReasoner as an interpretable elicitation instrument: a training-free approach that routes each step through modular neuro-cognitive lenses grounded in functional specificity and erotetic-reasoning theory. Across CogBench, math, and code evaluations, structured elicitation substantially improves vanilla inference and, in several domains, matches or exceeds post-trained thinking modes without parameter updates. At the same time, the framework exposes clear boundary conditions: risk-taking and uncertainty-sensitive cognitive control remain harder to recover through elicitation alone, and model scale modulates which signatures elicitation can reshape. These findings suggest that reasoning elicitation is not a uniform substitute for post-training, but a task- and scale-contingent mechanism that reveals both the strengths and limits of latent model reasoning. NeuReasoner offers an interpretable bridge between cognitive theory and inference-time model control, enabling systematic study of not only whether LLM reasoning can be improved, but which forms of reasoning can be elicited from contemporary reasoning models, under what conditions, and where the limits of elicitation lie.

\newpage
\bibliography{custom,anthology}
\bibliographystyle{colm2026_conference}

\clearpage
\appendix

\section{Compute-Matched Baselines}
\label{app:compute-matched}

NeuReasoner uses 11--39 LLM calls per decision; to test whether its gains
reflect the lens structure rather than inference-time compute alone, we
compare against self-consistency (SC) and iterative refinement (IR) matched
to NeuReasoner's per-task average call budget. SC draws $N$ independent
vanilla samples (temperature $0.7$ for diversity) \textbf{and takes a majority vote; IR runs $N$ sequential calls, each prepending the previous answer with a reconsideration prompt (\autoref{app:prompts})}. We note that SC requires non-zero temperature, a deliberate departure from the temperature-$0$ protocol used for NeuReasoner and the other CogBench conditions. All scores are normalized to random\,=\,0, human\,=\,1, matching \autoref{tab:summary_conditions}.
Primary metric is \texttt{performance\_score1}; cognitive signatures use
\texttt{behaviour\_score1}.

\begin{table}[t]
  \centering
  \caption{\textbf{Compute-matched performance, Qwen3-8B.} Normalized
  (random\,=\,0, human\,=\,1). $N$ = matched calls/decision. Scores $<0$ are
  below random. BART uses a single episode (feasibility); PR, HT, and TST are
  single deterministic engine-level statistics. Best per row in bold. Van.\,=\,Vanilla; NR\,=\,NeuReasoner.}
  \label{tab:cm-8b}
  \resizebox{\columnwidth}{!}{%
  \begin{tabular}{l c r r r r}
    \toprule
    \textbf{Experiment} & \textbf{N} & \textbf{Vanilla} & \textbf{SC} & \textbf{IR} & \textcolor{ourtext}{\textbf{NR}} \\
    \midrule
    Probabilistic Reasoning & 11 & $0.493$ & $0.908$ & $0.718$ & \cellcolor{oursrow}$\mathbf{1.570}$ \\
    Horizon Task (rew.)     & 11 & $-0.714$ & $0.726$ & $-0.661$ & \cellcolor{oursrow}$\mathbf{4.370}$ \\
    Restless Bandit (rew.)  & 23 & $0.152$ & $-0.009$ & $0.220$ & \cellcolor{oursrow}$\mathbf{0.850}$ \\
    Two-Step Task (rew.)    & 20 & $-0.004$ & $0.978$ & $\mathbf{1.247}$ & \cellcolor{oursrow}$0.442$ \\
    Instrumental Learning   & 18 & $0.643$ & $-0.274$ & $\mathbf{0.864}$ & \cellcolor{oursrow}$0.709$ \\
    BART (rew., 1 ep)       & 26 & $-0.001$ & $-0.039$ & $-0.034$ & \cellcolor{oursrow}$\mathbf{0.267}$ \\
    \bottomrule
  \end{tabular}}
\end{table}

\begin{table}[t]
  \centering
  \caption{\textbf{Compute-matched performance, Qwen3-32B.} The two 8B partial
  exceptions (Two-Step reward, Instrumental Learning) reverse: SC/IR fall to
  or below random while NeuReasoner leads. NR-8B shown for reference.}
  \label{tab:cm-32b}
  \resizebox{\columnwidth}{!}{%
  \begin{tabular}{l c r r r r r}
    \toprule
    \textbf{Experiment} & \textbf{N} & \textbf{Van.} & \textbf{SC} & \textbf{IR} & \textcolor{ourtext}{\textbf{NR}} & \textbf{NR-8B} \\
    \midrule
    Prob. Reasoning   & 17 & $1.021$ & $0.822$ & $-0.799$ & \cellcolor{oursrow}$\mathbf{1.599}$ & $1.570$ \\
    Horizon (rew.)    & 16 & $1.969$ & $0.075$ & $0.753$  & \cellcolor{oursrow}$\mathbf{3.263}$ & $4.370$ \\
    Restless B. (rew.)& 39 & $0.375$ & $0.309$ & $0.099$  & \cellcolor{oursrow}$\mathbf{0.857}$ & $0.850$ \\
    Two-Step (rew.)   & 33 & $-0.004$ & $0.963$ & $-2.160$ & \cellcolor{oursrow}$\mathbf{2.668}$ & $0.442$ \\
    Instr. Learning   & 17 & $0.952$ & $0.024$ & $-0.064$ & \cellcolor{oursrow}$\mathbf{1.880}$ & $0.709$ \\
    BART (rew., 1 ep) & 36 & $0.051$ & $0.042$ & $-0.002$ & \cellcolor{oursrow}$\mathbf{0.055}$ & $0.267$ \\
    \bottomrule
  \end{tabular}}
\end{table}

\begin{table}[t]
  \centering
  \caption{\textbf{Cognitive signatures} (\texttt{behaviour\_score1}),
  compute-matched. Only signatures where conditions diverge meaningfully are shown; experiments where all conditions cluster near the same value are omitted. NeuReasoner is the only positive condition on Restless
  Bandit meta-cognition at both scales. The Horizon directed-exploration
  signature is present at 8B but absent for all conditions at 32B.}
  \label{tab:cm-signatures}
  \resizebox{\columnwidth}{!}{%
  \begin{tabular}{l l r r r r}
    \toprule
    \textbf{Signature} & \textbf{Scale} & \textbf{Van.} & \textbf{SC} & \textbf{IR} & \textcolor{ourtext}{\textbf{NR}} \\
    \midrule
    HT directed expl.   & 8B  & $0.120$ & $0.120$ & $0.120$ & \cellcolor{oursrow}$\mathbf{1.358}$ \\
    HT directed expl.   & 32B & $-1.127$ & $-1.408$ & $-0.718$ & \cellcolor{oursrow}$\mathbf{-0.610}$ \\
    TST model-based     & 8B  & $0.837$ & $-0.084$ & $0.283$ & \cellcolor{oursrow}$\mathbf{4.409}$ \\
    TST model-based     & 32B & $3.046$ & $2.227$ & $0.800$ & \cellcolor{oursrow}$\mathbf{3.559}$ \\
    RB meta-cognition   & 8B  & $-1.898^{\dagger}$ & $-1.114$ & $-1.305$ & \cellcolor{oursrow}$\mathbf{0.109}$ \\
    RB meta-cognition   & 32B & $-1.898$ & $-1.462$ & $-1.305$ & \cellcolor{oursrow}$\mathbf{0.454}$ \\
    IL learning rate    & 32B & $0.636$ & $0.797$ & $0.046$ & \cellcolor{oursrow}$\mathbf{1.553}$ \\
    \bottomrule
  \end{tabular}}
  \\[2pt]
  {\footnotesize $^\dagger$~Vanilla RB meta-cognition baseline is identical at 32B.}

\end{table}

\section{Neural Lens Catalog}
\label{app:lenses}

This appendix gives the full description of the four neural lenses
introduced in \S\ref{sec:method}. Each lens is realized as a
prompt rubric executed in an isolated context by the same base model
under evaluation. The lens biases the paired inquiry operator's
analysis toward the kind of information the corresponding brain
network is associated with processing in humans; it does not act as a
standalone analyzer.

\paragraph{Language Network lens.}
Inspired by the cortical language network. Directs attention to the
linguistic surface of the question: wording, ambiguity, lexical scope,
framing, and notation. The orchestrator calls it when the question's
difficulty lies in how it is stated rather than in what it asks, for
example when a term is overloaded, when quantifier scope is unclear,
or when compact notation must be unpacked before any reasoning move
can be made on it.

\paragraph{Multiple-Demand lens.}
Inspired by the multiple-demand (frontoparietal control) network.
Directs attention to structure, constraints, controlled inference, and
goal maintenance under interference. The orchestrator calls it when
the question's difficulty is compositional or rule-governed: a chain
of constraints to be satisfied jointly, an inference that must track
several conditions at once, or a sub-goal that must be held in view
while another is pursued.

\paragraph{Theory-of-Mind lens.}
Inspired by the mentalizing network. Directs attention to agents,
beliefs, intentions, perspective, and strategic communication. The
orchestrator calls it when the question involves other minds: when an
answer depends on what an agent knows, intends, or would expect
another agent to do, or when the meaning of an utterance turns on the
speaker's communicative goal rather than its literal content.

\paragraph{Default Mode lens.}
Inspired by the default-mode network. Directs attention to simulation,
integration across loosely connected information, narrative coherence,
and self-projection. The orchestrator calls it when the question
requires imagining a trajectory rather than computing one: constructing
a plausible scenario, integrating background knowledge that is not
explicitly cued, or evaluating whether a proposed answer coheres with
a wider picture.

%
%
% ...

\section{Deferred Cognitive Inquiry Operators}
\label{app:inquiries}

This appendix gives the full description of the four inquiry operators. As with the lenses, each
operator is realized as a prompt rubric executed in an isolated context
by the same base model under evaluation. \textsc{Surface Issue} and
\textsc{Check Resolution}, which carry the framework's termination
logic, are listed in the main text; their full rubrics appear in Appendix~\ref{app:prompts}.

\paragraph{Expose Presuppositions.}
Lists the assumptions a candidate answer would inherit if pursued, and
flags the contestable ones for possible later rejection. The
orchestrator calls it when the live question rests on commitments that
have not yet been examined, or when an earlier \textsc{Check
Resolution} has failed in a way that suggests a presupposition rather
than the candidate itself is at fault.

\paragraph{Decompose Issue.}
Generates auxiliary subquestions and selects one as the next target.
Lets the agent recurse without losing the parent question, which
remains in scope at every step. Called when the live question is too
broad to attempt directly, or when a structuring move is needed before
any candidate can be proposed.

\paragraph{Pursue Answer.}
Proposes a candidate answer together with alternatives and supporting
evidence. The active answering move: the only operator that produces
a candidate, and the necessary precursor to \textsc{Check Resolution}.

\paragraph{Reopen Inquiry.}
Diagnoses the failure mode when \textsc{Check Resolution} returns
unresolved or partially resolved, and selects a revision strategy:
revise the issue (return to \textsc{Surface Issue}), reject a
presupposition (return to \textsc{Expose Presuppositions}), or
re-pursue with new evidence (return to \textsc{Pursue Answer}).
The operator names which prior step it is reopening, so the trace
records exactly what is being revised.

%
%
% ...

\definecolor{sysAccent}{HTML}{5F5E5A}    
\definecolor{neuroAccent}{HTML}{1F4E79}  
\definecolor{cogAccent}{HTML}{C0641E}    

\clearpage
\section{Prompts}
\label{app:prompts}

This appendix reproduces, verbatim, the prompt rubric for every operator
in the framework. Each operator is a specialized system prompt executed
by the same model under evaluation (Section~\ref{sec:framework-design});
no operator invokes external tools, code interpreters, or auxiliary
models. In the paper, we refer to the two operator families as
\emph{Neuro Lenses} and \emph{Cognitive Lenses}; in the raw prompts and
the runtime schema these correspond, respectively, to the
\texttt{neural\_lens} field and the \texttt{cog\_inquiry} field, and the
Cognitive Lens prompts retain their internal name ``inquiry operator''.
The orchestrator is governed by the system prompt below.

\subsection{System Prompt (Orchestrator)}
\label{app:system}

The orchestrator's system prompt is templated: the
\texttt{\{\% for \%\}} blocks are populated at runtime with the names
and descriptions of the available lenses. We reproduce the template
form.
\promptboxsys{System prompt (orchestrator)}{sysAccent}{prompts/system_v3.md}
\subsection{Neuro Lens Prompts}
\label{app:neuro-prompts}

\promptbox{Neuro Lens --- Language Network}{neuroAccent}{prompts/brain_lang_expert_v3.md}
\promptbox{Neuro Lens --- Multiple-Demand (Logic)}{neuroAccent}{prompts/brain_logic_expert_v3.md}
\promptbox{Neuro Lens --- Theory-of-Mind (Social)}{neuroAccent}{prompts/brain_social_reasoning_expert_v3.md}
\promptbox{Neuro Lens --- Default-Mode (World)}{neuroAccent}{prompts/brain_default_mode_v3.md}

\subsection{Cognitive Lens Prompts}
\label{app:cog-prompts}

\promptbox{Cognitive Lens --- Surface Issue}{cogAccent}{prompts/cog_surface_issue_v3.md}
\promptbox{Cognitive Lens --- Expose Presuppositions}{cogAccent}{prompts/cog_expose_presuppositions_v3.md}
\promptbox{Cognitive Lens --- Decompose Issue}{cogAccent}{prompts/cog_decompose_issue_v3.md}
\promptbox{Cognitive Lens --- Pursue Answer}{cogAccent}{prompts/cog_pursue_answer_v3.md}
\promptbox{Cognitive Lens --- Check Issue Resolution}{cogAccent}{prompts/cog_check_issue_resolution_v3.md}
\promptbox{Cognitive Lens --- Reopen Inquiry}{cogAccent}{prompts/cog_reopen_inquiry_v3.md}
\subsection{Python Coding Assistant}
\label{app:tool-prompts}
\promptbox{Python Coding Assistant}{toolAccent}{prompts/use_code_v3.md}

\section{Extended Results}
\label{app:results}

\subsection{Cognitive Phenotype Profiles}
\label{app:phenotype}

\begin{figure*}[t]
  \centering
  \includegraphics[width=\textwidth]{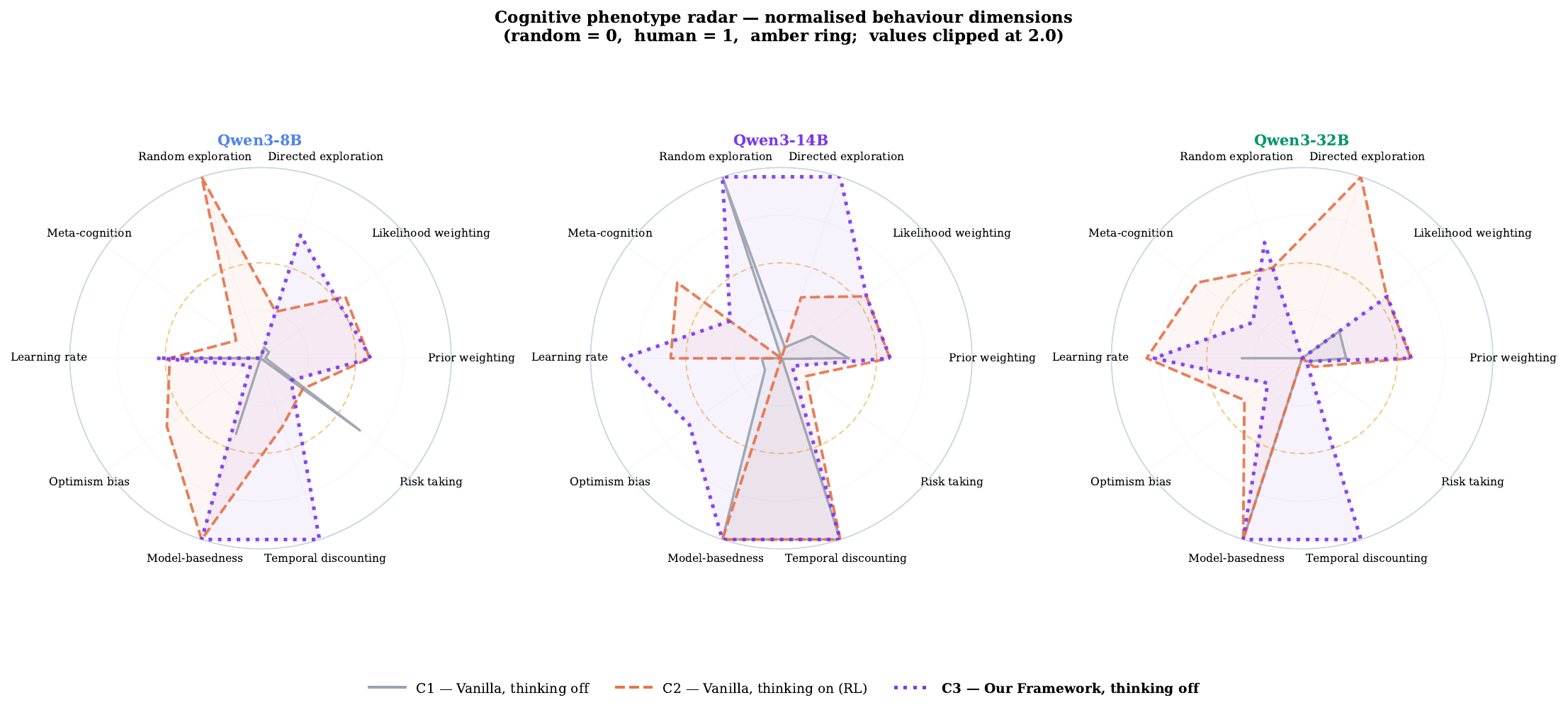}
  \caption{\textbf{Cognitive phenotype profiles --- Qwen3 family (C1\,/\,C2\,/\,C3).}
  Radar plots show all ten CogBench behavioral dimensions (random\,=\,0, human-average\,=\,1,
  amber ring) for Qwen3-\{8B,\,14B,\,32B\} under three conditions:
  \textbf{C1}~vanilla thinking off (solid grey), \textbf{C2}~RL-trained thinking on (dashed amber),
  and \textbf{C3}~NeuReasoner, thinking off (dotted green).
  C2 and C3 produce similar profiles on deliberation-heavy dimensions (Bayesian reasoning,
  model-based planning) but diverge markedly on exploration and heuristic dimensions,
  indicating that structured elicitation and RL post-training shape different facets of behavior.}
  \label{fig:exp3_phenotype_radar}
\end{figure*}

Figure~\ref{fig:exp3_phenotype_radar} shows radar profiles of all ten CogBench
behavioral dimensions for the Qwen3 family under all three conditions (C1/C2/C3).
Tables~\ref{tab:detailed_experiments} and~\ref{tab:phenotype_compact} give the
corresponding numerical values per experiment.



\begin{table}[t]
  \centering
    \caption{Per-experiment normalized performance (random\,$=0$, human\,$=1$).
    All conditions are fully complete (coverage matrix: 100\,\% in all cells).
    Values are mean\,$\pm$\,SEM across independently scored runs.
    For IL, RB, and BART each of the $N$ runs (8, 5, 5 respectively) is scored
    separately, yielding $N$ data points and a non-zero SEM.
    For PR, HT, and TST all collected episodes (10, 100, and 25 respectively) are
    pooled by the scorer into a single engine-level statistic---the scorer returns
    exactly one row per engine regardless of how many episodes were run---so
    SEM\,$=0$.
    For TD the task is not procedurally generated and runs once by design.
    \textcolor{ourtext}{\textbf{NeuReasoner}}\,=\,C3, thinking off.
    \textbf{Bold} = best score in each column within each experiment block.}
  \label{tab:detailed_experiments}
  \resizebox{\linewidth}{!}{%
  \begin{tabular}{l l r r r r}
    \toprule
    \textbf{Experiment} & \textbf{Condition} & \textbf{Qwen3-8B} & \textbf{Qwen3-14B} & \textbf{Qwen3-32B} & \textbf{Avg} \\
    \midrule
    \multirow{3}{*}{Bayesian Reasoning (PR)}
      & Baseline
        & $+0.493 \pm 0.000$ & $+0.967 \pm 0.000$ & $+1.021 \pm 0.000$ & $+0.827$ \\
      & Thinking ON (RL)
        & $\mathbf{+1.601 \pm 0.000}$ & $+1.598 \pm 0.000$ & $\mathbf{+1.602 \pm 0.000}$ & $\mathbf{+1.600}$ \\
    \rowcolor{oursrow}
      & \textcolor{ourtext}{\textbf{NeuReasoner}}
        & $+1.570 \pm 0.000$ & $\mathbf{+1.600 \pm 0.000}$ & $+1.599 \pm 0.000$ & $+1.590$ \\
    \arrayrulecolor{black}\cmidrule{2-6}
    \addlinespace[3pt]
    \multirow{3}{*}{Temporal Discounting (TD)}
      & Baseline
        & $-3.439 \pm 0.000$ & $\mathbf{+3.261 \pm 0.000}$ & $-4.276 \pm 0.000$ & $-1.485$ \\
      & Thinking ON (RL)
        & $+0.749 \pm 0.000$ & $+2.424 \pm 0.000$ & $-1.764 \pm 0.000$ & $+0.470$ \\
    \rowcolor{oursrow}
      & \textcolor{ourtext}{\textbf{NeuReasoner}}
        & $\mathbf{+2.424 \pm 0.000}$ & $+2.424 \pm 0.000$ & $\mathbf{+2.424 \pm 0.000}$ & $\mathbf{+2.424}$ \\
    \arrayrulecolor{black}\cmidrule{2-6}
    \addlinespace[3pt]
    \multirow{3}{*}{Directed Exploration (HT)}
      & Baseline
        & $-0.714 \pm 0.000$ & $\mathbf{+2.161 \pm 0.000}$ & $+1.969 \pm 0.000$ & $+1.139$ \\
      & Thinking ON (RL)
        & $+2.482 \pm 0.000$ & $+1.842 \pm 0.000$ & $-0.362 \pm 0.000$ & $+1.321$ \\
    \rowcolor{oursrow}
      & \textcolor{ourtext}{\textbf{NeuReasoner}}
        & $\mathbf{+4.370 \pm 0.000}$ & $+1.715 \pm 0.000$ & $\mathbf{+3.263 \pm 0.000}$ & $\mathbf{+3.116}$ \\
    \arrayrulecolor{black}\cmidrule{2-6}
    \addlinespace[3pt]
    \multirow{3}{*}{Reward Learning (IL)}
      & Baseline
        & $+0.643 \pm 0.552$ & $+0.798 \pm 0.486$ & $+0.952 \pm 0.345$ & $+0.798$ \\
      & Thinking ON (RL)
        & $\mathbf{+0.820 \pm 0.489}$ & $+0.820 \pm 0.369$ & $+0.709 \pm 0.291$ & $+0.783$ \\
    \rowcolor{oursrow}
      & \textcolor{ourtext}{\textbf{NeuReasoner}}
        & $+0.709 \pm 0.498$ & $\mathbf{+1.476 \pm 0.400}$ & $\mathbf{+1.880 \pm 0.341}$ & $\mathbf{+1.355}$ \\
    \arrayrulecolor{black}\cmidrule{2-6}
    \addlinespace[3pt]
    \multirow{3}{*}{Non-stationary Bandit (RB)}
      & Baseline
        & $+0.152 \pm 0.127$ & $+0.371 \pm 0.103$ & $+0.375 \pm 0.164$ & $+0.299$ \\
      & Thinking ON (RL)
        & $+0.612 \pm 0.195$ & $\mathbf{+1.358 \pm 0.071}$ & $\mathbf{+1.184 \pm 0.069}$ & $\mathbf{+1.051}$ \\
    \rowcolor{oursrow}
      & \textcolor{ourtext}{\textbf{NeuReasoner}}
        & $\mathbf{+0.850 \pm 0.182}$ & $+0.908 \pm 0.123$ & $+0.857 \pm 0.063$ & $+0.872$ \\
    \arrayrulecolor{black}\cmidrule{2-6}
    \addlinespace[3pt]
    \multirow{3}{*}{Risk-taking (BART)}
      & Baseline
        & $-0.001 \pm 0.035$ & $+0.012 \pm 0.018$ & $+0.051 \pm 0.006$ & $+0.021$ \\
      & Thinking ON (RL)
        & $+0.258 \pm 0.072$ & $\mathbf{+0.171 \pm 0.043}$ & $\mathbf{+0.134 \pm 0.010}$ & $\mathbf{+0.188}$ \\
    \rowcolor{oursrow}
      & \textcolor{ourtext}{\textbf{NeuReasoner}}
        & $\mathbf{+0.267 \pm 0.067}$ & $+0.111 \pm 0.014$ & $+0.055 \pm 0.008$ & $+0.144$ \\
    \arrayrulecolor{black}\cmidrule{2-6}
    \addlinespace[3pt]
    \multirow{3}{*}{Model-based RL (TST)}
      & Baseline
        & $-0.004 \pm 0.000$ & $+0.665 \pm 0.000$ & $-0.004 \pm 0.000$ & $+0.219$ \\
      & Thinking ON (RL)
        & $\mathbf{+0.665 \pm 0.000}$ & $\mathbf{+1.186 \pm 0.000}$ & $+1.260 \pm 0.000$ & $+1.037$ \\
    \rowcolor{oursrow}
      & \textcolor{ourtext}{\textbf{NeuReasoner}}
        & $+0.442 \pm 0.000$ & $+0.997 \pm 0.000$ & $\mathbf{+2.668 \pm 0.000}$ & $\mathbf{+1.369}$ \\
    \bottomrule
  \end{tabular}}
  \\[4pt]
  {\footnotesize \textit{SEM\,=\,0 for PR, HT, and TST because the CogBench scorer deliberately pools all trial-level responses from every session into a single regression, yielding one fitted coefficient per model--condition pair; one number has no variance to estimate. TD is not repeated by design. Experiments with non-zero SEM (IL, RB, BART) are scored independently per run, enabling within-experiment variance.}}
\end{table}

\begin{table*}[t]
  \centering
  \caption{Normalised behavioural phenotype dimensions (random\,$=0$, human\,$=1$). Experiments as in Table~\ref{tab:cogbench}. Three conditions: \textbf{C1}\,=\,vanilla thinking off; \textbf{C2}\,=\,vanilla RL-trained thinking on; \colorbox{Orchid!10}{\strut\textbf{C3}}\,=\,NeuReasoner (thinking off). \textcolor{OliveGreen}{Green}\,$\geq\!1.0$; neutral $0$--$1$; \textcolor{BrickRed}{red}\,$<\!0$. \textbf{Bold}\,=\,best condition per model size.}
  \label{tab:phenotype_compact}
  \resizebox{\linewidth}{!}{%
  \begin{tabular}{ll rrr rrr rrr}
    \toprule
    \textbf{Experiment} & \textbf{Dimension} & \multicolumn{3}{c}{\textbf{Qwen3-8B}} & \multicolumn{3}{c}{\textbf{Qwen3-14B}} & \multicolumn{3}{c}{\textbf{Qwen3-32B}} \\
    \arrayrulecolor{black}\cmidrule(lr){3-5}\cmidrule(lr){6-8}\cmidrule(lr){9-11}
    & & C1 & C2 & \cellcolor{Orchid!10}\textbf{C3} & C1 & C2 & \cellcolor{Orchid!10}\textbf{C3} & C1 & C2 & \cellcolor{Orchid!10}\textbf{C3} \\
    \arrayrulecolor{black}\midrule
    \multirow{2}{*}{PR} & Prior weighting    & 0.05 & \textcolor{OliveGreen}{1.14} & \cellcolor{Orchid!10}\textbf{\textcolor{OliveGreen}{1.15}} & 0.71 & \textcolor{OliveGreen}{1.14} & \cellcolor{Orchid!10}\textbf{\textcolor{OliveGreen}{1.14}} & 0.46 & \textcolor{OliveGreen}{1.14} & \cellcolor{Orchid!10}\textbf{\textcolor{OliveGreen}{1.15}} \\
                        & Likelihood weighting & 0.11 & \textbf{\textcolor{OliveGreen}{1.10}} & \cellcolor{Orchid!10}\textcolor{OliveGreen}{1.01} & 0.40 & \textcolor{OliveGreen}{1.10} & \cellcolor{Orchid!10}\textbf{\textcolor{OliveGreen}{1.10}} & 0.48 & \textcolor{OliveGreen}{1.10} & \cellcolor{Orchid!10}\textbf{\textcolor{OliveGreen}{1.10}} \\
    \midrule
    \multirow{2}{*}{HT} & Directed exploration & 0.12 & 0.51 & \cellcolor{Orchid!10}\textbf{\textcolor{OliveGreen}{1.36}} & 0.12 & 0.67 & \cellcolor{Orchid!10}\textbf{\textcolor{OliveGreen}{2.67}} & \textcolor{BrickRed}{-1.13} & \textbf{\textcolor{OliveGreen}{3.18}} & \cellcolor{Orchid!10}\textcolor{BrickRed}{-0.61} \\
                        & Random exploration  & \textcolor{BrickRed}{-0.91} & \textbf{\textcolor{OliveGreen}{7.66}} & \cellcolor{Orchid!10}\textcolor{BrickRed}{-2.19} & \textcolor{OliveGreen}{3.00} & \textcolor{BrickRed}{-0.06} & \cellcolor{Orchid!10}\textbf{\textcolor{OliveGreen}{4.97}} & \textcolor{BrickRed}{-1.45} & 1.00 & \cellcolor{Orchid!10}\textbf{\textcolor{OliveGreen}{1.28}} \\
    \midrule
    RB                  & Meta-cognition      & \textcolor{BrickRed}{-1.86} & \textbf{0.32} & \cellcolor{Orchid!10}\textcolor{BrickRed}{-0.24} & \textcolor{BrickRed}{-1.96} & \textbf{\textcolor{OliveGreen}{1.35}} & \cellcolor{Orchid!10}0.66 & \textcolor{BrickRed}{-2.63} & \textbf{\textcolor{OliveGreen}{1.35}} & \cellcolor{Orchid!10}0.64 \\
    \midrule
    \multirow{2}{*}{IL} & Learning rate      & 0.83 & 0.95 & \cellcolor{Orchid!10}\textbf{\textcolor{OliveGreen}{1.09}} & 0.21 & \textcolor{OliveGreen}{1.16} & \cellcolor{Orchid!10}\textbf{\textcolor{OliveGreen}{1.66}} & 0.64 & \textbf{\textcolor{OliveGreen}{1.63}} & \cellcolor{Orchid!10}\textcolor{OliveGreen}{1.55} \\
                        & Optimism bias       & \textcolor{BrickRed}{-0.62} & \textbf{\textcolor{OliveGreen}{1.21}} & \cellcolor{Orchid!10}0.13 & 0.21 & \textcolor{BrickRed}{-1.97} & \cellcolor{Orchid!10}\textbf{\textcolor{OliveGreen}{1.19}} & \textcolor{BrickRed}{-0.18} & \textbf{0.75} & \cellcolor{Orchid!10}0.45 \\
    \midrule
    TST                 & Model-basedness     & 0.84 & \textbf{\textcolor{OliveGreen}{5.38}} & \cellcolor{Orchid!10}\textcolor{OliveGreen}{4.41} & \textcolor{OliveGreen}{2.20} & \textbf{\textcolor{OliveGreen}{6.13}} & \cellcolor{Orchid!10}\textcolor{OliveGreen}{5.53} & \textcolor{OliveGreen}{3.05} & \textbf{\textcolor{OliveGreen}{6.77}} & \cellcolor{Orchid!10}\textcolor{OliveGreen}{3.56} \\
    \midrule
    TD                  & Temporal discounting & \textcolor{BrickRed}{-3.44} & 0.75 & \cellcolor{Orchid!10}\textbf{\textcolor{OliveGreen}{2.42}} & \textbf{\textcolor{OliveGreen}{3.26}} & \textcolor{OliveGreen}{2.42} & \cellcolor{Orchid!10}\textcolor{OliveGreen}{2.42} & \textcolor{BrickRed}{-4.28} & \textcolor{BrickRed}{-1.76} & \cellcolor{Orchid!10}\textbf{\textcolor{OliveGreen}{2.42}} \\
    \midrule
    BART                & Risk taking         & \textbf{\textcolor{OliveGreen}{1.29}} & 0.55 & \cellcolor{Orchid!10}0.39 & 0.01 & \textbf{0.32} & \cellcolor{Orchid!10}0.14 & 0.06 & \textbf{0.15} & \cellcolor{Orchid!10}0.06 \\
    \arrayrulecolor{black}\bottomrule
  \end{tabular}}
\end{table*}

\subsection{Math and Code Benchmarks}
\label{app:math-code}

\begin{figure*}[t]
  \centering
  \includegraphics[width=\textwidth]{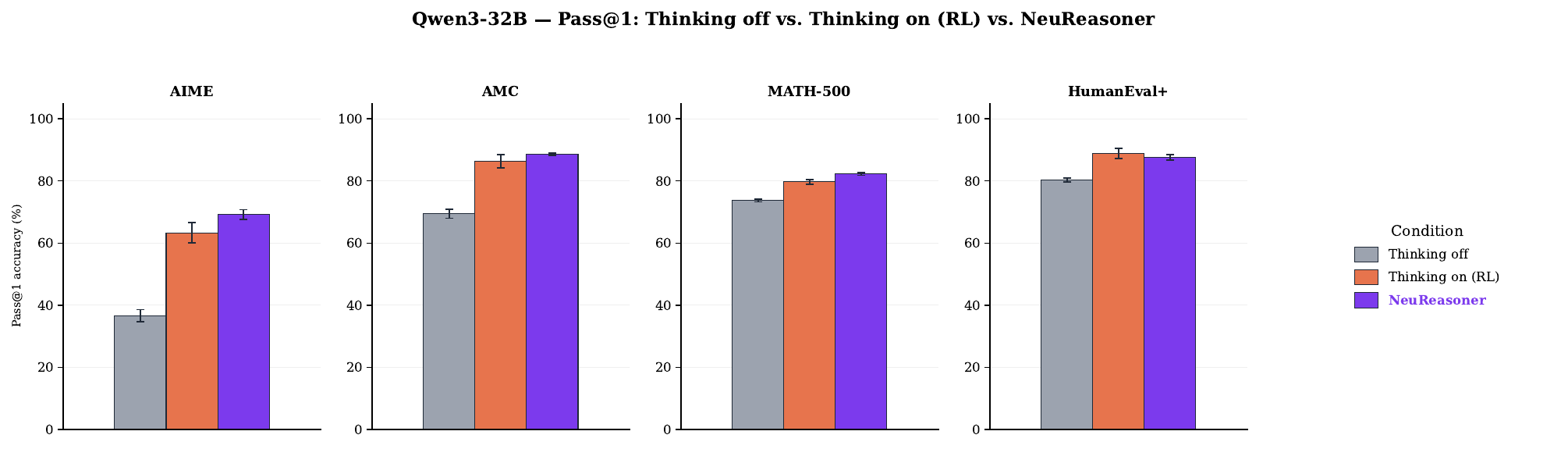}
  \caption{\textbf{Math and code benchmark results --- Qwen3-32B (C1\,/\,C2\,/\,C3).}
  Grouped bars show Pass@1 accuracy (\%) across four tasks for three conditions:
  \textbf{C1}~Thinking off (\textcolor[HTML]{9CA3AF}{$\blacksquare$}~grey, vanilla),
  \textbf{C2}~Thinking on\textsuperscript{RL} (\textcolor[HTML]{E7744D}{$\blacksquare$}~orange,
  RL-trained), and
  \textbf{C3}~NeuReasoner (\textcolor[HTML]{7C3AED}{$\blacksquare$}~purple, thinking off).
  Error bars show $\pm$\,SEM across $K\!=\!3$--$4$ repetitions.
  NeuReasoner leads on MATH-500 (82.2\,\% vs.\ 79.7\,\%) and matches thinking-on on AMC (88.6\,\%
  vs.\ 86.3\,\%), while trailing on HumanEval\texttt{+} by a narrow margin (87.6\,\% vs.\ 88.9\,\%).
  Full numerical results for all models are in Table~\ref{tab:math_human_eval}.}
  \label{fig:math_threeway_app}
\end{figure*}

Figure~\ref{fig:math_threeway_app} and Table~\ref{tab:math_human_eval} report Pass@1 accuracy
on math reasoning (AIME, AMC, MATH-500) and code generation (HumanEval\texttt{+}) for
Qwen3-8B and Qwen3-32B across the three conditions.
NeuReasoner matches or exceeds the thinking-on baseline for Qwen3-32B on three of four tasks
(winning outright on AIME, AMC, and MATH-500), but trails substantially on Qwen3-8B
where RL-trained chain-of-thought already reaches ceiling-level performance on AIME and AMC.

\begin{table*}[t]
  \centering
  \setlength{\tabcolsep}{5pt}
  \renewcommand{\arraystretch}{1.2}
  \caption{Pass@1 accuracy (\%) on math reasoning (AIME, AMC, MATH-500) and code generation (HumanEval\texttt{+}), averaged across $K\!=\!3$--$8$ independent repetitions per condition. Three conditions are compared within the same model: \textit{Thinking off} (vanilla chain-of-thought, no scaffold), \textit{Thinking on\textsuperscript{RL}} (native RL-trained reasoning mode), and \textbf{NeuReasoner} (our cognitive scaffold, thinking off). Values shown as mean${\scriptstyle\,\pm\,}$SEM. \cellcolor{mygreen!25}\textbf{Highlighted} = best result per column. Bottom row: NeuReasoner $\Delta$ vs.\ best available baseline (\textcolor{mygreen}{\textbf{green}}\,=\,gain; \textcolor{myred}{\textbf{red}}\,=\,deficit; shading\,$\propto$\,$|\Delta|$).}
  \label{tab:math_human_eval}
  \resizebox{\linewidth}{!}{
  \begin{tabular}{l |rrrr |rrrr}
    \toprule
     & \multicolumn{4}{c}{\textbf{Qwen3-8B}} & \multicolumn{4}{c}{\textbf{Qwen3-32B}} \\
    \cmidrule(lr){2-5} \cmidrule(lr){6-9}
    \textbf{Condition} & AIME & AMC & M-500 & HE\texttt{+} & AIME & AMC & M-500 & HE\texttt{+} \\
    \midrule
    \rowcolor{black!5}    Thinking off & $25.6{\scriptstyle\,\pm\,1.11}$ & $59.8{\scriptstyle\,\pm\,0.80}$ & $72.3{\scriptstyle\,\pm\,0.07}$ & $78.5{\scriptstyle\,\pm\,0.52}$ & $36.7{\scriptstyle\,\pm\,1.92}$ & $69.5{\scriptstyle\,\pm\,1.45}$ & $73.7{\scriptstyle\,\pm\,0.41}$ & $80.3{\scriptstyle\,\pm\,0.68}$ \\
    \rowcolor{Melon!20}    Thinking on\textsuperscript{RL} & \cellcolor{mygreen!25}$\mathbf{74.4{\scriptstyle\,\pm\,2.22}}$ & \cellcolor{mygreen!25}$\mathbf{87.6{\scriptstyle\,\pm\,1.06}}$ & \cellcolor{mygreen!25}$\mathbf{80.1{\scriptstyle\,\pm\,0.24}}$ & \cellcolor{mygreen!25}$\mathbf{87.8{\scriptstyle\,\pm\,0.66}}$ & $63.3{\scriptstyle\,\pm\,3.33}$ & $86.3{\scriptstyle\,\pm\,2.13}$ & $79.7{\scriptstyle\,\pm\,0.75}$ & \cellcolor{mygreen!25}$\mathbf{88.9{\scriptstyle\,\pm\,1.64}}$ \\
    \rowcolor{Orchid!15}    \textcolor{ourtext}{\textbf{NeuReasoner}} & $29.2{\scriptstyle\,\pm\,4.79}$ & $50.3{\scriptstyle\,\pm\,2.93}$ & $68.4{\scriptstyle\,\pm\,0.37}$ & $79.7{\scriptstyle\,\pm\,1.24}$ & \cellcolor{mygreen!25}$\mathbf{69.2{\scriptstyle\,\pm\,1.60}}$ & \cellcolor{mygreen!25}$\mathbf{88.6{\scriptstyle\,\pm\,0.35}}$ & \cellcolor{mygreen!25}$\mathbf{82.2{\scriptstyle\,\pm\,0.46}}$ & $87.6{\scriptstyle\,\pm\,0.81}$ \\
    \midrule
    \textit{NeuReasoner\;$\Delta$} & \cellcolor{myred!45}$-45.3$ & \cellcolor{myred!37}$-37.2$ & \cellcolor{myred!12}$-11.7$ & \cellcolor{myred!8}$-8.1$ & \cellcolor{mygreen!6}$+5.8$ & \cellcolor{mygreen!2}$+2.2$ & \cellcolor{mygreen!3}$+2.5$ & \cellcolor{myred!1}$-1.3$ \\
    \bottomrule
  \end{tabular}
  }
\end{table*}

\section{Lens-Selection Analysis}
\label{app:lens-analysis}

\subsection{Brain-Lens Usage}
\label{app:brain-lens}

\begin{figure*}[t]
  \centering
  \includegraphics[width=\textwidth]{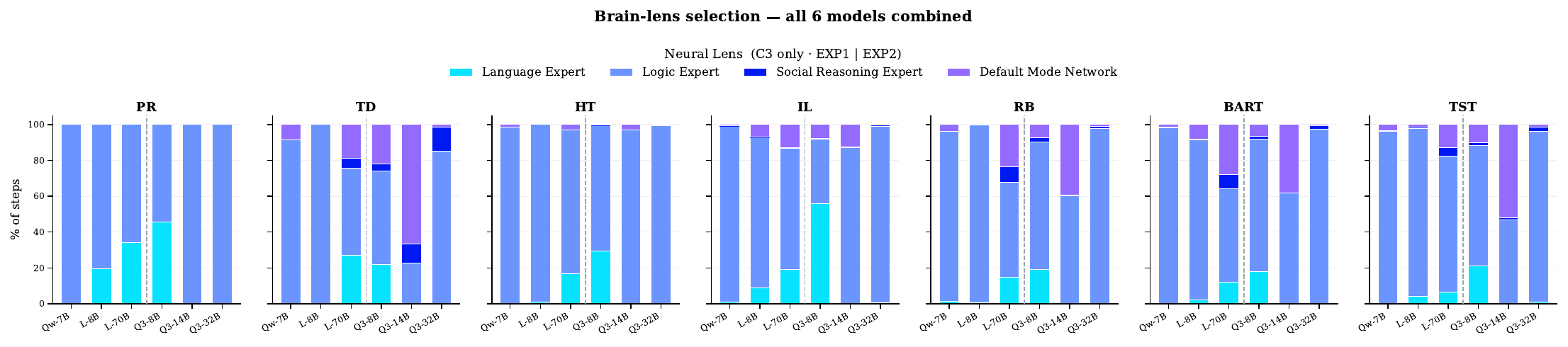}
  \caption{\textbf{Brain-lens selection --- Qwen3 family, all experiments.}
  Stacked bars show the percentage of reasoning steps allocated to each Neuro Lens
  (\textbf{Lang}=Language Network, \textbf{MD}=Multiple-Demand,
  \textbf{ToM}=Theory-of-Mind, \textbf{DMN}=Default-Mode Network),
  pooled across corresponding models for each CogBench task.
  The Multiple-Demand lens dominates in all seven experiments.
  Per-model breakdowns are in Table~\ref{tab:brain_lens_per_model}.}
  \label{fig:brain_lens_combined}
\end{figure*}

Figure~\ref{fig:brain_lens_combined} shows brain-lens selection pooled across the given models.
Table~\ref{tab:brain_lens_per_model} gives per-model, per-experiment breakdowns.

\begin{figure}[t]
  \centering
  \includegraphics[width=0.72\textwidth]{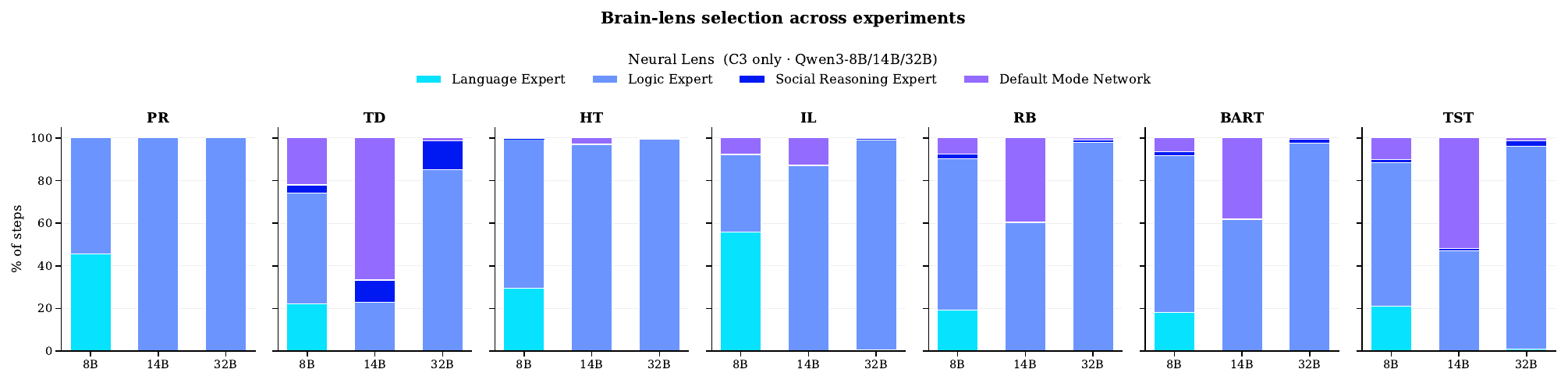}
  \caption{\textbf{Brain $\times$ Cognitive lens selection summary} averaged across models for each CogBench experiment.}
  \label{fig:lens_heatmap_exp1-body}
\end{figure}

\begin{table}[t]\centering
  \caption{Brain-lens selection by model and experiment (NeuReasoner (C3), \% of reasoning steps). Color intensity proportional to usage frequency. \textbf{Lang}=Language Expert, \textbf{MD}=Logic Expert, \textbf{ToM}=Social Reasoning, \textbf{DMN}=Default Mode Network.}
  \label{tab:brain_lens_per_model}
  \resizebox{\linewidth}{!}{%
  \begin{tabular}{l rrr r rrr r rrr r }
    \toprule
    \textbf{Exp.} & \multicolumn{4}{c}{\textbf{Qwen3-8B}} & \multicolumn{4}{c}{\textbf{Qwen3-14B}} & \multicolumn{4}{c}{\textbf{Qwen3-32B}} \\
    \arrayrulecolor{black}\cmidrule(lr){2-5} \cmidrule(lr){6-9} \cmidrule(lr){10-13}
    & \textit{Lang} & \textit{MD} & \textit{ToM} & \textit{DMN} & \textit{Lang} & \textit{MD} & \textit{ToM} & \textit{DMN} & \textit{Lang} & \textit{MD} & \textit{ToM} & \textit{DMN} \\
    \arrayrulecolor{black}\midrule
    PR & \cellcolor{RoyalBlue!18}45.8 & \cellcolor{RoyalBlue!22}\textbf{54.2} & 0.0 & 0.0 & 0.0 & \cellcolor{RoyalBlue!40}\textbf{100.0} & 0.0 & 0.0 & 0.0 & \cellcolor{RoyalBlue!40}\textbf{100.0} & 0.0 & 0.0 \\
    TD & \cellcolor{RoyalBlue!9}22.0 & \cellcolor{RoyalBlue!21}\textbf{52.0} & 4.0 & \cellcolor{RoyalBlue!9}22.0 & 0.0 & \cellcolor{RoyalBlue!9}22.8 & \cellcolor{RoyalBlue!4}10.5 & \cellcolor{RoyalBlue!27}\textbf{66.7} & 0.0 & \cellcolor{RoyalBlue!34}\textbf{85.2} & \cellcolor{RoyalBlue!5}13.3 & 1.5 \\
    HT & \cellcolor{RoyalBlue!12}29.5 & \cellcolor{RoyalBlue!28}\textbf{70.2} & 0.1 & 0.3 & 0.0 & \cellcolor{RoyalBlue!39}\textbf{97.0} & 0.1 & 2.9 & 0.0 & \cellcolor{RoyalBlue!40}\textbf{99.3} & 0.6 & 0.1 \\
    IL & \cellcolor{RoyalBlue!22}\textbf{55.8} & \cellcolor{RoyalBlue!14}36.2 & 0.5 & 7.6 & 0.0 & \cellcolor{RoyalBlue!35}\textbf{87.1} & 0.4 & \cellcolor{RoyalBlue!5}12.6 & 0.7 & \cellcolor{RoyalBlue!39}\textbf{98.2} & 0.7 & 0.3 \\
    RB & \cellcolor{RoyalBlue!8}19.1 & \cellcolor{RoyalBlue!29}\textbf{71.3} & 2.1 & 7.5 & 0.0 & \cellcolor{RoyalBlue!24}\textbf{60.2} & 0.4 & \cellcolor{RoyalBlue!16}39.5 & 0.3 & \cellcolor{RoyalBlue!39}\textbf{97.6} & 1.2 & 0.9 \\
    BART & \cellcolor{RoyalBlue!7}18.1 & \cellcolor{RoyalBlue!29}\textbf{73.6} & 1.7 & 6.5 & 0.0 & \cellcolor{RoyalBlue!25}\textbf{61.8} & 0.2 & \cellcolor{RoyalBlue!15}38.0 & 0.3 & \cellcolor{RoyalBlue!39}\textbf{97.3} & 1.8 & 0.6 \\
    TST & \cellcolor{RoyalBlue!8}21.2 & \cellcolor{RoyalBlue!27}\textbf{67.2} & 1.6 & \cellcolor{RoyalBlue!4}10.0 & 0.0 & \cellcolor{RoyalBlue!19}46.9 & 1.3 & \cellcolor{RoyalBlue!21}\textbf{51.8} & 0.8 & \cellcolor{RoyalBlue!38}\textbf{95.4} & 2.4 & 1.4 \\
    \arrayrulecolor{black}\bottomrule
  \end{tabular}}
\end{table}

\subsection{Cognitive-Inquiry Operator Usage}
\label{app:cog-inquiry}

\begin{figure*}[t]
  \centering
  \includegraphics[width=\textwidth]{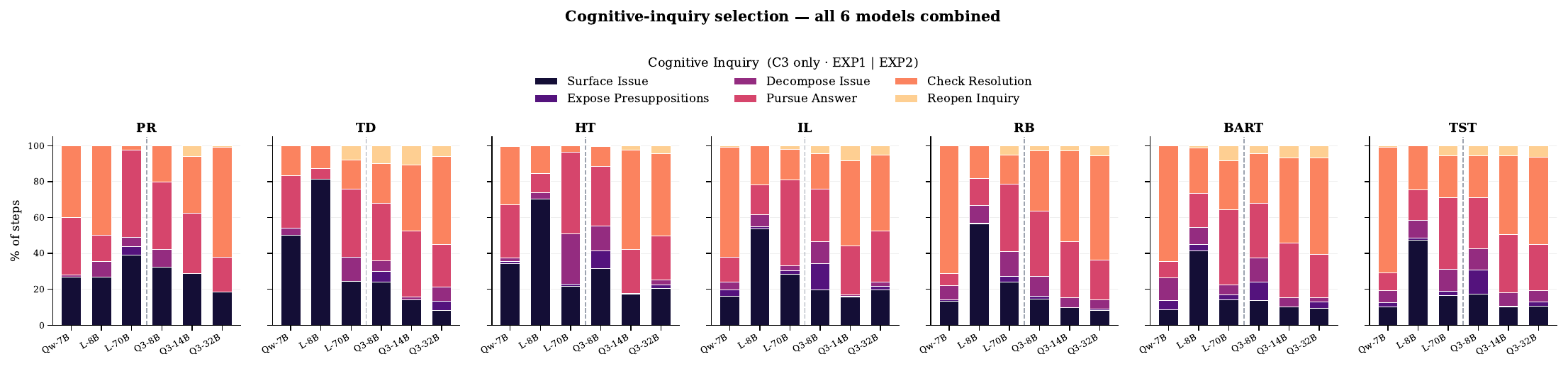}
  \caption{\textbf{Cognitive-inquiry operator selection, all experiments.}
  Stacked bars show the percentage of steps assigned to each Cognitive Lens
  (\textbf{Surface}=Surface Issue, \textbf{Expose}=Expose Presuppositions,
  \textbf{Dec.}=Decompose Issue, \textbf{Pursue}=Pursue Answer,
  \textbf{Check}=Check Resolution, \textbf{Reopen}=Reopen Inquiry),
  pooled across corresponding models.
  Pursue Answer and Check Resolution together account for the majority of steps,
  reflecting a strong preference for direct answer-seeking over exploratory or diagnostic moves.
  Per-model breakdowns are in Table~\ref{tab:cog_inquiry_per_model}.}
  \label{fig:cog_inquiry_combined}
\end{figure*}

Figure~\ref{fig:cog_inquiry_combined} shows cognitive-inquiry operator selection pooled across the given models. Table~\ref{tab:cog_inquiry_per_model} gives per-model breakdowns.

\begin{table}[t]\centering
  \caption{Cognitive-inquiry operator selection by model and experiment (NeuReasoner (C3), \% of reasoning steps). Color intensity proportional to usage. \textbf{Surface}=Surface Issue, \textbf{Expose}=Expose Presuppositions, \textbf{Dec.}=Decompose Issue, \textbf{Pursue}=Pursue Answer, \textbf{Check}=Check Resolution, \textbf{Reopen}=Reopen Inquiry.}
  \label{tab:cog_inquiry_per_model}
  \resizebox{\linewidth}{!}{%
  \begin{tabular}{l rrrrrr rrrrrr rrrrrr }
    \toprule
    \textbf{Exp.} & \multicolumn{6}{c}{\textbf{Qwen3-8B}} & \multicolumn{6}{c}{\textbf{Qwen3-14B}} & \multicolumn{6}{c}{\textbf{Qwen3-32B}} \\
    \arrayrulecolor{black}\cmidrule(lr){2-7} \cmidrule(lr){8-13} \cmidrule(lr){14-19}
    & \textit{Surface} & \textit{Expose} & \textit{Decompose} & \textit{Pursue} & \textit{Check} & \textit{Reopen} & \textit{Surface} & \textit{Expose} & \textit{Decompose} & \textit{Pursue} & \textit{Check} & \textit{Reopen} & \textit{Surface} & \textit{Expose} & \textit{Decompose} & \textit{Pursue} & \textit{Check} & \textit{Reopen} \\
    \arrayrulecolor{black}\midrule
    PR & \cellcolor{BurntOrange!21}32.2 & 0.0 & \cellcolor{BurntOrange!7}10.2 & \cellcolor{BurntOrange!24}\textbf{37.3} & \cellcolor{BurntOrange!13}20.3 & 0.0 & \cellcolor{BurntOrange!19}29.0 & 0.0 & 0.0 & \cellcolor{BurntOrange!22}\textbf{33.3} & \cellcolor{BurntOrange!21}31.9 & \cellcolor{BurntOrange!4}5.8 & \cellcolor{BurntOrange!12}18.5 & 0.0 & 0.0 & \cellcolor{BurntOrange!13}19.4 & \cellcolor{BurntOrange!40}\textbf{61.1} & 0.9 \\
    TD & \cellcolor{BurntOrange!16}24.0 & \cellcolor{BurntOrange!4}6.0 & \cellcolor{BurntOrange!4}6.0 & \cellcolor{BurntOrange!21}\textbf{32.0} & \cellcolor{BurntOrange!14}22.0 & \cellcolor{BurntOrange!7}10.0 & \cellcolor{BurntOrange!9}14.0 & 0.0 & 1.8 & \cellcolor{BurntOrange!24}\textbf{36.8} & \cellcolor{BurntOrange!24}\textbf{36.8} & \cellcolor{BurntOrange!7}10.5 & \cellcolor{BurntOrange!5}8.1 & 5.2 & \cellcolor{BurntOrange!5}8.1 & \cellcolor{BurntOrange!16}23.7 & \cellcolor{BurntOrange!32}\textbf{48.9} & \cellcolor{BurntOrange!4}5.9 \\
    HT & \cellcolor{BurntOrange!21}31.6 & \cellcolor{BurntOrange!6}9.7 & \cellcolor{BurntOrange!9}14.0 & \cellcolor{BurntOrange!22}\textbf{33.0} & \cellcolor{BurntOrange!7}11.4 & 0.3 & \cellcolor{BurntOrange!11}17.3 & 0.0 & 0.3 & \cellcolor{BurntOrange!16}24.8 & \cellcolor{BurntOrange!36}\textbf{55.0} & 2.6 & \cellcolor{BurntOrange!13}20.6 & 1.8 & 2.8 & \cellcolor{BurntOrange!16}24.4 & \cellcolor{BurntOrange!30}\textbf{45.8} & 4.6 \\
    IL & \cellcolor{BurntOrange!13}19.8 & \cellcolor{BurntOrange!9}14.3 & \cellcolor{BurntOrange!8}12.5 & \cellcolor{BurntOrange!19}\textbf{29.0} & \cellcolor{BurntOrange!13}20.1 & 4.3 & \cellcolor{BurntOrange!10}15.7 & 0.6 & 0.7 & \cellcolor{BurntOrange!18}27.1 & \cellcolor{BurntOrange!31}\textbf{47.6} & \cellcolor{BurntOrange!5}8.2 & \cellcolor{BurntOrange!13}19.7 & 2.0 & 2.3 & \cellcolor{BurntOrange!19}28.7 & \cellcolor{BurntOrange!27}\textbf{42.0} & \cellcolor{BurntOrange!4}5.4 \\
    RB & \cellcolor{BurntOrange!9}14.5 & 1.5 & \cellcolor{BurntOrange!7}11.1 & \cellcolor{BurntOrange!24}\textbf{36.6} & \cellcolor{BurntOrange!22}33.7 & 2.6 & \cellcolor{BurntOrange!6}9.9 & 0.0 & \cellcolor{BurntOrange!4}5.5 & \cellcolor{BurntOrange!21}31.4 & \cellcolor{BurntOrange!33}\textbf{50.3} & 3.0 & \cellcolor{BurntOrange!5}8.4 & 0.8 & 4.9 & \cellcolor{BurntOrange!15}22.3 & \cellcolor{BurntOrange!38}\textbf{58.2} & \cellcolor{BurntOrange!4}5.4 \\
    BART & \cellcolor{BurntOrange!9}13.7 & \cellcolor{BurntOrange!7}10.3 & \cellcolor{BurntOrange!9}13.5 & \cellcolor{BurntOrange!20}\textbf{30.4} & \cellcolor{BurntOrange!18}27.8 & 4.4 & \cellcolor{BurntOrange!7}10.3 & 0.1 & 4.8 & \cellcolor{BurntOrange!20}30.7 & \cellcolor{BurntOrange!31}\textbf{47.3} & \cellcolor{BurntOrange!4}6.7 & \cellcolor{BurntOrange!6}9.4 & 3.5 & 2.6 & \cellcolor{BurntOrange!16}23.9 & \cellcolor{BurntOrange!35}\textbf{53.8} & \cellcolor{BurntOrange!4}6.8 \\
    TST & \cellcolor{BurntOrange!11}17.5 & \cellcolor{BurntOrange!9}13.4 & \cellcolor{BurntOrange!8}11.7 & \cellcolor{BurntOrange!19}\textbf{28.4} & \cellcolor{BurntOrange!15}23.5 & \cellcolor{BurntOrange!4}5.5 & \cellcolor{BurntOrange!7}10.4 & 0.3 & \cellcolor{BurntOrange!5}7.5 & \cellcolor{BurntOrange!21}32.3 & \cellcolor{BurntOrange!29}\textbf{43.9} & \cellcolor{BurntOrange!4}5.7 & \cellcolor{BurntOrange!7}10.6 & 2.2 & \cellcolor{BurntOrange!4}6.7 & \cellcolor{BurntOrange!17}25.6 & \cellcolor{BurntOrange!32}\textbf{48.6} & \cellcolor{BurntOrange!4}6.3 \\
    \arrayrulecolor{black}\bottomrule
  \end{tabular}}
\end{table}

\subsection{Lens \texorpdfstring{$\times$}{x} Inquiry Co-occurrence}
\label{app:lens-heatmaps}

\begin{figure*}[t]
  \centering
  \includegraphics[width=\textwidth]{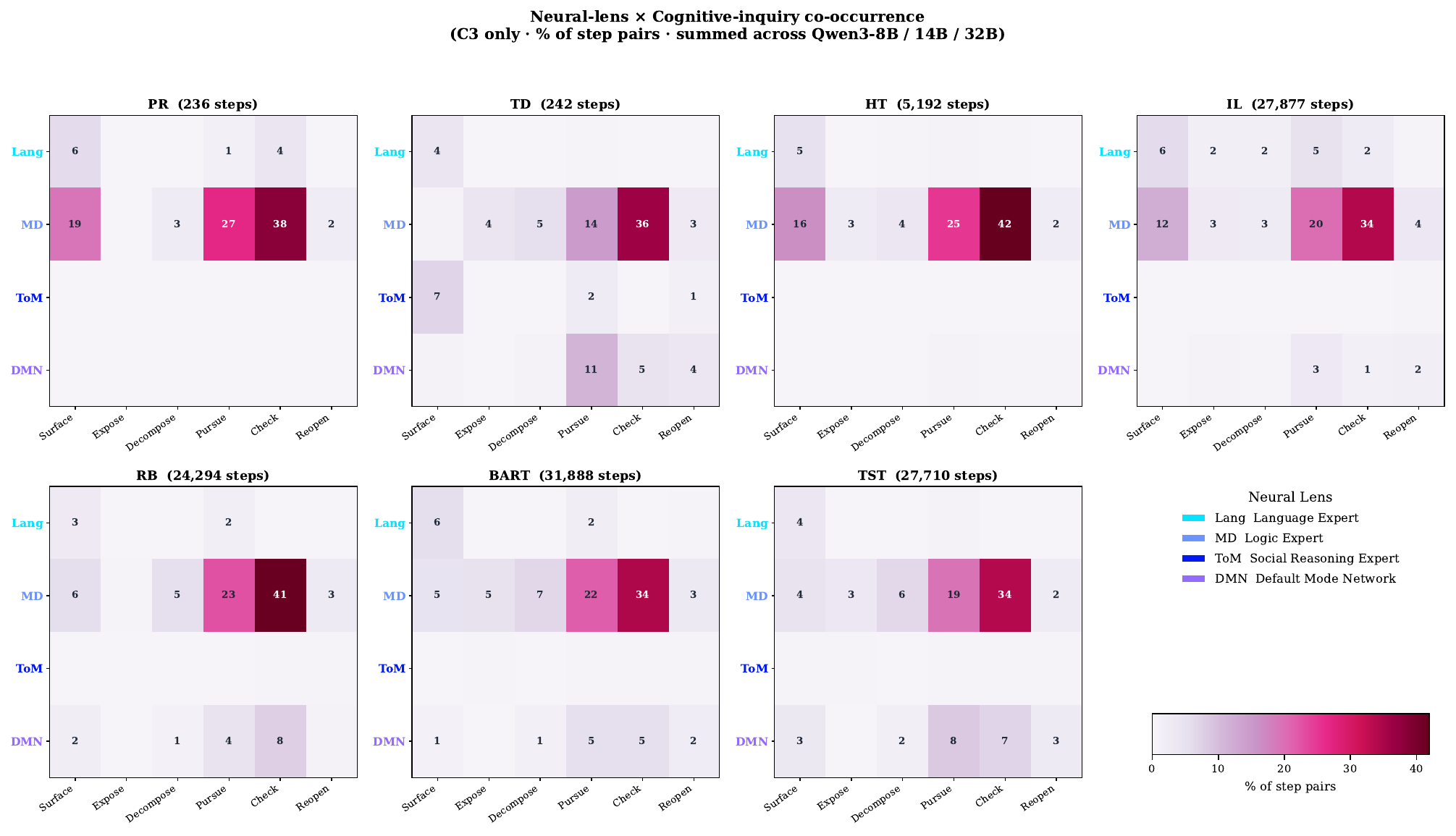}
  \caption{\textbf{Brain-lens $\times$ cognitive-inquiry co-occurrence --- per model and task.}
  Each heatmap shows the percentage of reasoning steps pairing each (Neuro Lens, Cognitive Lens)
  combination, broken down by Qwen3 model and CogBench experiment.
  Cell values $\geq$\,1\% are annotated.
  Even at this per-model resolution, the mass concentrates in one or two cells per panel,
  confirming a catalog-collapse that is consistent across model sizes.}
  \label{fig:lens_heatmap_exp1}
\end{figure*}

\begin{figure*}[t]
  \centering
  \includegraphics[width=\textwidth]{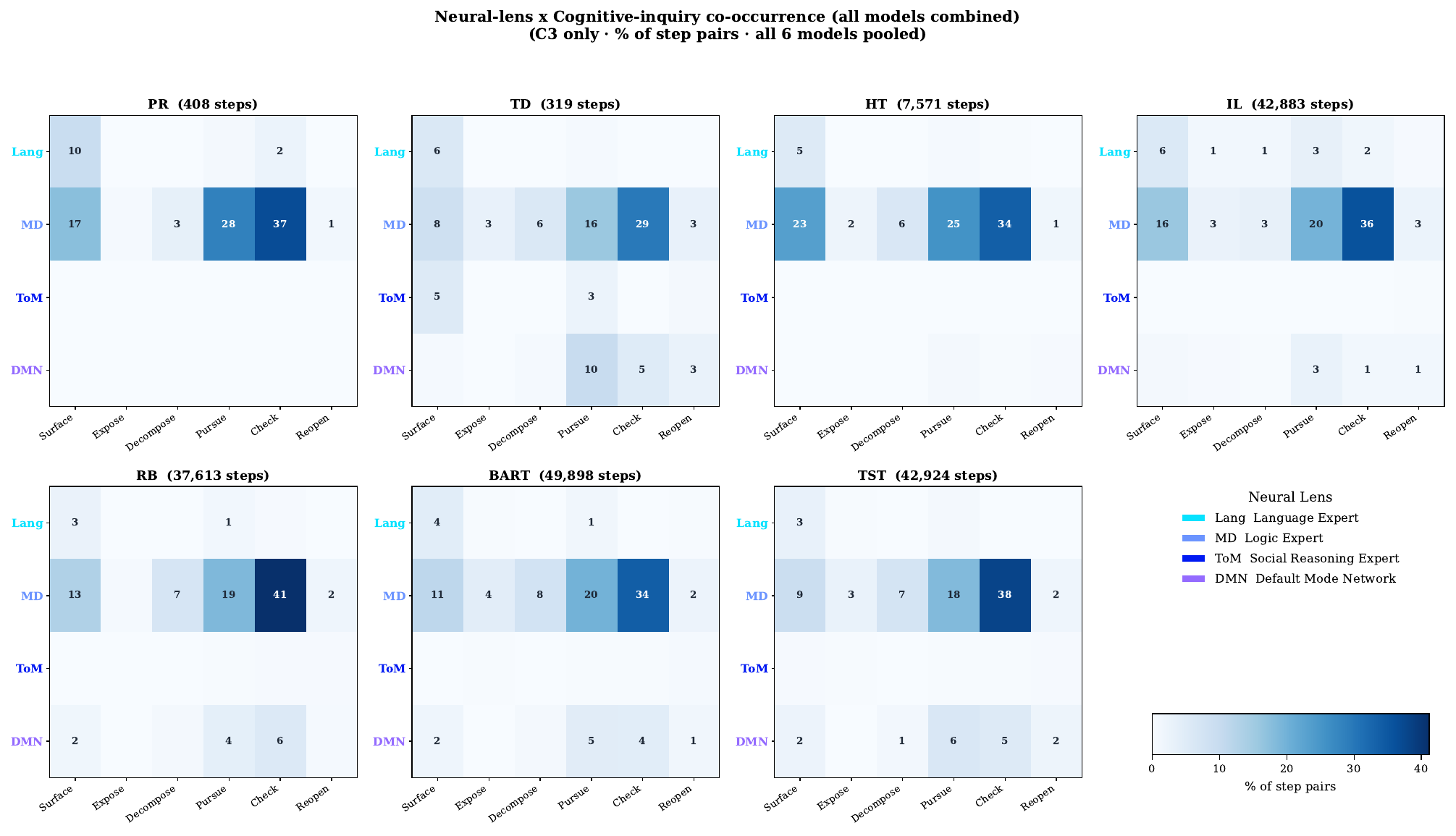}
  \caption{\textbf{Brain-lens $\times$ cognitive-inquiry co-occurrence --- pooled, Qwen3 family.}
  Heatmaps aggregate pair-selection frequencies across Qwen3-\{8B,\,14B,\,32B\} and all experiments.
  The dominant Multiple-Demand $\times$ Pursue-Answer pairing persists at the aggregate level,
  while most of the 4\,$\times$\,6 catalog remains near-zero.
  Compare with Figure~\ref{fig:lens_heatmap_exp1} for the per-model view.}
  \label{fig:lens_heatmap_combined}
\end{figure*}

Figures~\ref{fig:lens_heatmap_exp1}--\ref{fig:lens_heatmap_combined} show (Neuro Lens, Cognitive Lens)
pair selection frequencies per model and pooled across the Qwen3 family respectively.
Table~\ref{tab:cog_neural_per_task_model} gives the full numerical breakdown per model and experiment.

\begin{sidewaystable}[p]\centering
  \caption{Neural-lens $\times$ cognitive-inquiry co-occurrence per experiment and model (NeuReasoner (C3), \% of steps; rows normalized within each neural lens). Color intensity reflects pairing frequency. Brain lenses: \textbf{Lang}=Language Expert, \textbf{MD}=Logic Expert, \textbf{ToM}=Social Reasoning, \textbf{DMN}=Default Mode Network.}
  \label{tab:cog_neural_per_task_model}
  \resizebox{\linewidth}{!}{%
  \begin{tabular}{l l @{\hspace{4pt}} rrrrrr @{\hspace{4pt}} rrrrrr @{\hspace{4pt}} rrrrrr}
    \toprule
    \textbf{Exp.} & \textbf{Lens} & \multicolumn{6}{c}{\textbf{Qwen3-8B}} & \multicolumn{6}{c}{\textbf{Qwen3-14B}} & \multicolumn{6}{c}{\textbf{Qwen3-32B}} \\
    \arrayrulecolor{black}\cmidrule(lr){3-8} \cmidrule(lr){9-14} \cmidrule(lr){15-20}
    & & \textit{Sur} & \textit{Exp} & \textit{Dec} & \textit{Pur} & \textit{Che} & \textit{Reo} & \textit{Sur} & \textit{Exp} & \textit{Dec} & \textit{Pur} & \textit{Che} & \textit{Reo} & \textit{Sur} & \textit{Exp} & \textit{Dec} & \textit{Pur} & \textit{Che} & \textit{Reo} \\
    \arrayrulecolor{black}\midrule
    \multirow{4}{*}{\textbf{PR}} & \textcolor{RoyalBlue}{Lang} & \cellcolor{RoyalBlue!21}\textbf{51.9} & 0.0 & 0.0 & \cellcolor{RoyalBlue!4}11.1 & \cellcolor{RoyalBlue!15}37.0 & 0.0 & 0.0 & 0.0 & 0.0 & 0.0 & 0.0 & 0.0 & 0.0 & 0.0 & 0.0 & 0.0 & 0.0 & 0.0 \\
     & \textcolor{Plum}{MD} & \cellcolor{RoyalBlue!6}15.6 & 0.0 & \cellcolor{RoyalBlue!8}18.8 & \cellcolor{RoyalBlue!24}\textbf{59.4} & 6.2 & 0.0 & \cellcolor{RoyalBlue!12}29.0 & 0.0 & 0.0 & \cellcolor{RoyalBlue!13}\textbf{33.3} & \cellcolor{RoyalBlue!13}31.9 & 5.8 & \cellcolor{RoyalBlue!7}18.5 & 0.0 & 0.0 & \cellcolor{RoyalBlue!8}19.4 & \cellcolor{RoyalBlue!24}\textbf{61.1} & 0.9 \\
     & \textcolor{OliveGreen}{ToM} & 0.0 & 0.0 & 0.0 & 0.0 & 0.0 & 0.0 & 0.0 & 0.0 & 0.0 & 0.0 & 0.0 & 0.0 & 0.0 & 0.0 & 0.0 & 0.0 & 0.0 & 0.0 \\
     & \textcolor{Goldenrod}{DMN} & 0.0 & 0.0 & 0.0 & 0.0 & 0.0 & 0.0 & 0.0 & 0.0 & 0.0 & 0.0 & 0.0 & 0.0 & 0.0 & 0.0 & 0.0 & 0.0 & 0.0 & 0.0 \\
    \midrule
    \multirow{4}{*}{\textbf{TD}} & \textcolor{RoyalBlue}{Lang} & \cellcolor{RoyalBlue!36}\textbf{90.9} & 0.0 & 0.0 & \cellcolor{RoyalBlue!4}9.1 & 0.0 & 0.0 & 0.0 & 0.0 & 0.0 & 0.0 & 0.0 & 0.0 & 0.0 & 0.0 & 0.0 & 0.0 & 0.0 & 0.0 \\
     & \textcolor{Plum}{MD} & 7.7 & \cellcolor{RoyalBlue!5}11.5 & \cellcolor{RoyalBlue!5}11.5 & \cellcolor{RoyalBlue!9}23.1 & \cellcolor{RoyalBlue!15}\textbf{38.5} & 7.7 & 0.0 & 0.0 & 0.0 & 7.7 & \cellcolor{RoyalBlue!37}\textbf{92.3} & 0.0 & 0.0 & 6.1 & 8.7 & \cellcolor{RoyalBlue!10}24.3 & \cellcolor{RoyalBlue!23}\textbf{56.5} & 4.3 \\
     & \textcolor{OliveGreen}{ToM} & 0.0 & 0.0 & 0.0 & \cellcolor{RoyalBlue!40}\textbf{100.0} & 0.0 & 0.0 & \cellcolor{RoyalBlue!40}\textbf{100.0} & 0.0 & 0.0 & 0.0 & 0.0 & 0.0 & \cellcolor{RoyalBlue!24}\textbf{61.1} & 0.0 & 0.0 & \cellcolor{RoyalBlue!9}22.2 & 0.0 & \cellcolor{RoyalBlue!7}16.7 \\
     & \textcolor{Goldenrod}{DMN} & 0.0 & 0.0 & 0.0 & \cellcolor{RoyalBlue!25}\textbf{63.6} & \cellcolor{RoyalBlue!4}9.1 & \cellcolor{RoyalBlue!11}27.3 & 5.3 & 0.0 & 2.6 & \cellcolor{RoyalBlue!21}\textbf{52.6} & \cellcolor{RoyalBlue!9}23.7 & \cellcolor{RoyalBlue!6}15.8 & 0.0 & 0.0 & \cellcolor{RoyalBlue!20}\textbf{50.0} & 0.0 & \cellcolor{RoyalBlue!20}\textbf{50.0} & 0.0 \\
    \midrule
    \multirow{4}{*}{\textbf{HT}} & \textcolor{RoyalBlue}{Lang} & \cellcolor{RoyalBlue!32}\textbf{79.2} & 2.3 & 3.2 & \cellcolor{RoyalBlue!5}11.7 & 3.5 & 0.0 & 0.0 & 0.0 & 0.0 & 0.0 & 0.0 & 0.0 & 0.0 & 0.0 & 0.0 & 0.0 & 0.0 & 0.0 \\
     & \textcolor{Plum}{MD} & \cellcolor{RoyalBlue!5}11.7 & \cellcolor{RoyalBlue!5}12.8 & \cellcolor{RoyalBlue!7}18.6 & \cellcolor{RoyalBlue!17}\textbf{41.8} & \cellcolor{RoyalBlue!6}14.8 & 0.2 & \cellcolor{RoyalBlue!7}17.8 & 0.0 & 0.3 & \cellcolor{RoyalBlue!10}24.2 & \cellcolor{RoyalBlue!22}\textbf{56.2} & 1.5 & \cellcolor{RoyalBlue!8}20.7 & 1.9 & 2.7 & \cellcolor{RoyalBlue!10}24.5 & \cellcolor{RoyalBlue!18}\textbf{46.0} & 4.2 \\
     & \textcolor{OliveGreen}{ToM} & 0.0 & 0.0 & 0.0 & 0.0 & 0.0 & \cellcolor{RoyalBlue!40}\textbf{100.0} & 0.0 & \cellcolor{RoyalBlue!20}\textbf{50.0} & 0.0 & \cellcolor{RoyalBlue!20}\textbf{50.0} & 0.0 & 0.0 & 0.0 & 0.0 & 0.0 & \cellcolor{RoyalBlue!9}22.2 & \cellcolor{RoyalBlue!4}11.1 & \cellcolor{RoyalBlue!27}\textbf{66.7} \\
     & \textcolor{Goldenrod}{DMN} & 0.0 & 0.0 & 0.0 & \cellcolor{RoyalBlue!40}\textbf{100.0} & 0.0 & 0.0 & 0.0 & 0.0 & 0.0 & \cellcolor{RoyalBlue!17}\textbf{43.7} & \cellcolor{RoyalBlue!7}18.3 & \cellcolor{RoyalBlue!15}38.0 & 0.0 & 0.0 & \cellcolor{RoyalBlue!40}\textbf{100.0} & 0.0 & 0.0 & 0.0 \\
    \midrule
    \multirow{4}{*}{\textbf{IL}} & \textcolor{RoyalBlue}{Lang} & \cellcolor{RoyalBlue!14}\textbf{35.4} & \cellcolor{RoyalBlue!4}9.1 & \cellcolor{RoyalBlue!4}10.2 & \cellcolor{RoyalBlue!11}28.0 & \cellcolor{RoyalBlue!5}13.6 & 3.8 & 0.0 & 0.0 & 0.0 & 0.0 & 0.0 & 0.0 & 5.3 & \cellcolor{RoyalBlue!6}15.8 & 0.0 & \cellcolor{RoyalBlue!26}\textbf{64.9} & \cellcolor{RoyalBlue!5}12.3 & 1.8 \\
     & \textcolor{Plum}{MD} & 0.1 & \cellcolor{RoyalBlue!8}20.3 & \cellcolor{RoyalBlue!7}16.9 & \cellcolor{RoyalBlue!12}29.5 & \cellcolor{RoyalBlue!13}\textbf{32.3} & 0.8 & \cellcolor{RoyalBlue!7}17.9 & 0.3 & 0.4 & \cellcolor{RoyalBlue!10}24.4 & \cellcolor{RoyalBlue!21}\textbf{51.3} & 5.8 & \cellcolor{RoyalBlue!8}20.0 & 1.7 & 2.3 & \cellcolor{RoyalBlue!11}28.4 & \cellcolor{RoyalBlue!17}\textbf{42.6} & 5.0 \\
     & \textcolor{OliveGreen}{ToM} & 2.5 & 2.5 & 5.0 & \cellcolor{RoyalBlue!4}10.0 & 7.5 & \cellcolor{RoyalBlue!29}\textbf{72.5} & 0.0 & \cellcolor{RoyalBlue!10}23.8 & \cellcolor{RoyalBlue!6}14.3 & \cellcolor{RoyalBlue!6}14.3 & 4.8 & \cellcolor{RoyalBlue!17}\textbf{42.9} & 1.8 & \cellcolor{RoyalBlue!10}24.6 & 0.0 & \cellcolor{RoyalBlue!5}12.3 & \cellcolor{RoyalBlue!4}8.8 & \cellcolor{RoyalBlue!21}\textbf{52.6} \\
     & \textcolor{Goldenrod}{DMN} & 0.3 & \cellcolor{RoyalBlue!10}25.4 & 8.7 & \cellcolor{RoyalBlue!14}\textbf{35.5} & \cellcolor{RoyalBlue!4}9.8 & \cellcolor{RoyalBlue!8}20.2 & 1.1 & 1.9 & 2.8 & \cellcolor{RoyalBlue!18}\textbf{46.2} & \cellcolor{RoyalBlue!10}23.8 & \cellcolor{RoyalBlue!10}24.3 & 0.0 & \cellcolor{RoyalBlue!7}17.9 & \cellcolor{RoyalBlue!7}17.9 & \cellcolor{RoyalBlue!20}\textbf{50.0} & 0.0 & \cellcolor{RoyalBlue!6}14.3 \\
    \midrule
    \multirow{4}{*}{\textbf{RB}} & \textcolor{RoyalBlue}{Lang} & \cellcolor{RoyalBlue!25}\textbf{63.0} & 0.1 & 0.1 & \cellcolor{RoyalBlue!14}34.6 & 1.5 & 0.7 & 0.0 & 0.0 & 0.0 & 0.0 & 0.0 & 0.0 & 0.0 & 6.2 & 0.0 & \cellcolor{RoyalBlue!9}21.9 & \cellcolor{RoyalBlue!22}\textbf{56.2} & \cellcolor{RoyalBlue!6}15.6 \\
     & \textcolor{Plum}{MD} & 3.5 & 2.0 & \cellcolor{RoyalBlue!6}15.2 & \cellcolor{RoyalBlue!16}\textbf{39.3} & \cellcolor{RoyalBlue!15}38.2 & 1.9 & 7.4 & 0.0 & 5.3 & \cellcolor{RoyalBlue!13}32.8 & \cellcolor{RoyalBlue!21}\textbf{53.1} & 1.5 & 8.6 & 0.5 & 4.4 & \cellcolor{RoyalBlue!9}22.4 & \cellcolor{RoyalBlue!24}\textbf{59.0} & 5.1 \\
     & \textcolor{OliveGreen}{ToM} & 0.0 & 1.6 & 0.0 & \cellcolor{RoyalBlue!4}9.5 & \cellcolor{RoyalBlue!19}\textbf{47.6} & \cellcolor{RoyalBlue!17}41.3 & 0.0 & 0.0 & \cellcolor{RoyalBlue!9}23.3 & \cellcolor{RoyalBlue!8}20.0 & \cellcolor{RoyalBlue!12}\textbf{30.0} & \cellcolor{RoyalBlue!11}26.7 & 1.7 & \cellcolor{RoyalBlue!9}21.8 & 7.6 & \cellcolor{RoyalBlue!6}14.3 & \cellcolor{RoyalBlue!10}25.2 & \cellcolor{RoyalBlue!12}\textbf{29.4} \\
     & \textcolor{Goldenrod}{DMN} & 0.5 & 0.2 & 3.8 & \cellcolor{RoyalBlue!9}23.1 & \cellcolor{RoyalBlue!28}\textbf{68.8} & 3.6 & \cellcolor{RoyalBlue!6}13.8 & 0.0 & 5.7 & \cellcolor{RoyalBlue!12}29.4 & \cellcolor{RoyalBlue!18}\textbf{46.1} & 4.9 & 0.0 & 6.9 & \cellcolor{RoyalBlue!21}\textbf{52.9} & \cellcolor{RoyalBlue!6}14.9 & \cellcolor{RoyalBlue!7}18.4 & 6.9 \\
    \midrule
    \multirow{4}{*}{\textbf{BART}} & \textcolor{RoyalBlue}{Lang} & \cellcolor{RoyalBlue!28}\textbf{71.2} & 1.8 & 0.6 & \cellcolor{RoyalBlue!9}22.9 & 0.8 & 2.8 & 0.0 & 0.0 & 0.0 & \cellcolor{RoyalBlue!40}\textbf{100.0} & 0.0 & 0.0 & 0.0 & \cellcolor{RoyalBlue!7}16.7 & 0.0 & \cellcolor{RoyalBlue!20}\textbf{50.0} & \cellcolor{RoyalBlue!13}33.3 & 0.0 \\
     & \textcolor{Plum}{MD} & 1.0 & \cellcolor{RoyalBlue!5}12.6 & \cellcolor{RoyalBlue!6}15.9 & \cellcolor{RoyalBlue!12}31.2 & \cellcolor{RoyalBlue!15}\textbf{36.6} & 2.7 & \cellcolor{RoyalBlue!4}11.0 & 0.1 & 4.7 & \cellcolor{RoyalBlue!12}29.7 & \cellcolor{RoyalBlue!20}\textbf{50.8} & 3.7 & \cellcolor{RoyalBlue!4}9.6 & 3.1 & 2.4 & \cellcolor{RoyalBlue!10}24.0 & \cellcolor{RoyalBlue!22}\textbf{54.5} & 6.4 \\
     & \textcolor{OliveGreen}{ToM} & 1.2 & \cellcolor{RoyalBlue!13}31.8 & 1.2 & \cellcolor{RoyalBlue!11}26.5 & 0.8 & \cellcolor{RoyalBlue!15}\textbf{38.4} & \cellcolor{RoyalBlue!4}10.0 & \cellcolor{RoyalBlue!4}10.0 & 0.0 & \cellcolor{RoyalBlue!8}20.0 & \cellcolor{RoyalBlue!8}20.0 & \cellcolor{RoyalBlue!16}\textbf{40.0} & 3.1 & \cellcolor{RoyalBlue!8}19.1 & 3.8 & \cellcolor{RoyalBlue!6}15.3 & \cellcolor{RoyalBlue!15}\textbf{36.6} & \cellcolor{RoyalBlue!9}22.1 \\
     & \textcolor{Goldenrod}{DMN} & 0.1 & 2.8 & \cellcolor{RoyalBlue!10}24.6 & \cellcolor{RoyalBlue!17}\textbf{43.0} & \cellcolor{RoyalBlue!4}11.0 & \cellcolor{RoyalBlue!7}18.4 & \cellcolor{RoyalBlue!4}9.2 & 0.2 & 5.2 & \cellcolor{RoyalBlue!13}32.5 & \cellcolor{RoyalBlue!17}\textbf{41.6} & \cellcolor{RoyalBlue!5}11.4 & 2.1 & 8.3 & \cellcolor{RoyalBlue!14}\textbf{35.4} & \cellcolor{RoyalBlue!8}18.8 & \cellcolor{RoyalBlue!5}12.5 & \cellcolor{RoyalBlue!9}22.9 \\
    \midrule
    \multirow{4}{*}{\textbf{TST}} & \textcolor{RoyalBlue}{Lang} & \cellcolor{RoyalBlue!32}\textbf{79.3} & 2.9 & 0.5 & \cellcolor{RoyalBlue!5}13.1 & 1.5 & 2.7 & 0.0 & 0.0 & 0.0 & 0.0 & 0.0 & 0.0 & 2.4 & 7.1 & 0.0 & \cellcolor{RoyalBlue!25}\textbf{63.5} & \cellcolor{RoyalBlue!9}22.4 & 4.7 \\
     & \textcolor{Plum}{MD} & 0.5 & \cellcolor{RoyalBlue!7}18.4 & \cellcolor{RoyalBlue!6}14.9 & \cellcolor{RoyalBlue!12}30.9 & \cellcolor{RoyalBlue!13}\textbf{32.8} & 2.5 & 2.6 & 0.2 & \cellcolor{RoyalBlue!4}10.2 & \cellcolor{RoyalBlue!12}29.1 & \cellcolor{RoyalBlue!23}\textbf{56.9} & 0.9 & \cellcolor{RoyalBlue!4}10.7 & 1.8 & 6.4 & \cellcolor{RoyalBlue!10}25.7 & \cellcolor{RoyalBlue!20}\textbf{50.2} & 5.3 \\
     & \textcolor{OliveGreen}{ToM} & \cellcolor{RoyalBlue!7}18.4 & 8.2 & 3.1 & \cellcolor{RoyalBlue!5}13.3 & 5.1 & \cellcolor{RoyalBlue!21}\textbf{52.0} & \cellcolor{RoyalBlue!20}\textbf{49.6} & 2.8 & 6.4 & \cellcolor{RoyalBlue!4}10.6 & \cellcolor{RoyalBlue!5}12.8 & \cellcolor{RoyalBlue!7}17.7 & \cellcolor{RoyalBlue!5}13.5 & \cellcolor{RoyalBlue!7}18.1 & 2.3 & \cellcolor{RoyalBlue!4}10.8 & \cellcolor{RoyalBlue!8}20.4 & \cellcolor{RoyalBlue!14}\textbf{35.0} \\
     & \textcolor{Goldenrod}{DMN} & 0.2 & 2.4 & \cellcolor{RoyalBlue!6}15.7 & \cellcolor{RoyalBlue!19}\textbf{46.6} & \cellcolor{RoyalBlue!4}10.4 & \cellcolor{RoyalBlue!10}24.8 & \cellcolor{RoyalBlue!7}16.4 & 0.3 & 5.0 & \cellcolor{RoyalBlue!14}\textbf{35.6} & \cellcolor{RoyalBlue!13}32.9 & \cellcolor{RoyalBlue!4}9.7 & 6.1 & 2.0 & \cellcolor{RoyalBlue!16}\textbf{38.8} & \cellcolor{RoyalBlue!8}20.4 & 6.8 & \cellcolor{RoyalBlue!10}25.9 \\
    \arrayrulecolor{black}\bottomrule
  \end{tabular}}
  \smallskip\par\noindent{\footnotesize Abbreviations: Sur=Surface, Exp=Expose, Dec=Decompose, Pur=Pursue, Chk=Check, Reo=Reopen.}
\end{sidewaystable}

\subsection{Computational Cost}
\label{app:cost}

\begin{figure*}[t]
  \centering
  \includegraphics[width=\textwidth]{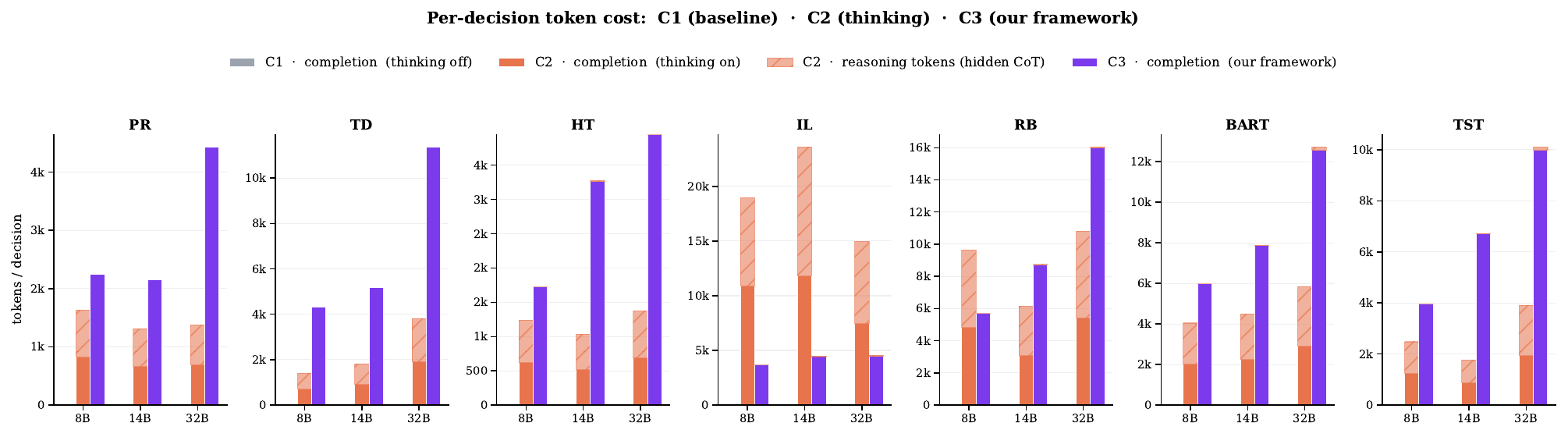}
  \caption{\textbf{Per-decision token cost across conditions.}
  Output tokens per decision for C1 (vanilla, thinking off), C2 (thinking on, RL-trained),
  and C3 (NeuReasoner, thinking off), broken down by model and task.
  C2 hidden chain-of-thought tokens (reasoning) are shown separately from completion tokens.
  C3 incurs higher total token counts than C1 due to multiple lens calls per decision,
  but remains substantially cheaper per call than C2's RL-trained reasoning traces;
  prompt caching on fixed lens system prompts (${\approx}47\%$ of C3 input) would reduce C3 cost by 40--50\%.}
  \label{fig:token_cost}
\end{figure*}

Figure~\ref{fig:token_cost} visualizes per-decision token cost across conditions.
Tables~\ref{tab:token-cost}--\ref{tab:token-cost-per-task} give per-model and per-task numerical breakdowns.

\begin{table}[t]
  \centering
  \caption{Mean output tokens per decision, LLM call count, and estimated inference cost (averaged across all seven tasks). \textit{Comp}\,=\,completion tokens; \textit{Reas}\,=\,hidden chain-of-thought (\mbox{C2} thinking mode only); \textbf{Total}\,=\,Comp\,+\,Reas; \textbf{Calls}\,=\,mean LLM calls per decision (\mbox{C3} call count varies 12--34 across tasks depending on task complexity); \textbf{Cost/k}\,=\,estimated USD per 1,000 decisions (OpenRouter rates, May~2026: Qwen3-8B~\$0.05/\$0.40, 14B~\$0.10/\$0.24, 32B~\$0.08/\$0.28 per\,M\,tokens in/out; reasoning tokens billed as output). Per LLM call, \mbox{C3} costs \$0.27--0.49/k vs. \$1.57--2.28/k for \mbox{C2} (5--12\,\texttimes{} cheaper per call); \mbox{C3} total cost scales with reasoning depth, not per-call inefficiency. Enabling prompt caching for fixed tool system prompts (${\approx}47\%$ of \mbox{C3} input) would reduce \mbox{C3} costs by 40--50\%.}
  \label{tab:token-cost}
  \resizebox{\linewidth}{!}{
  \begin{tabular}{l |rrrrr |rrrrr |rrrrr}
    \toprule
    & \multicolumn{5}{c}{\textbf{Qwen3-8B}} & \multicolumn{5}{c}{\textbf{Qwen3-14B}} & \multicolumn{5}{c}{\textbf{Qwen3-32B}} \\
    \cmidrule(lr){2-6} \cmidrule(lr){7-11} \cmidrule(lr){12-16}
    Condition & Comp. & Reas. & Total & Calls & Cost/k & Comp. & Reas. & Total & Calls & Cost/k & Comp. & Reas. & Total & Calls & Cost/k \\
    \midrule
    \rowcolor{black!5}C1  (thinking off) & 1 & --- & \textbf{1} & \textcolor{gray}{1} & \textbf{\$0.029} & 2 & --- & \textbf{2} & \textcolor{gray}{1} & \textbf{\$0.057} & 2 & --- & \textbf{2} & \textcolor{gray}{1} & \textbf{\$0.045} \\
    \rowcolor{Melon!20}C2  (thinking on) & 3.0k & \textit{\textbf{2.6k}} & \textbf{5.6k} & \textcolor{gray}{1} & \textbf{\$2.28} & 2.9k & \textit{\textbf{2.9k}} & \textbf{5.7k} & \textcolor{gray}{1} & \textbf{\$1.57} & 3.0k & \textit{\textbf{3.0k}} & \textbf{6.0k} & \textcolor{gray}{1} & \textbf{\$1.73} \\
    \rowcolor{Orchid!15}NeuReasoner (C3) & 3.9k & --- & \textbf{3.9k} & \textbf{\textcolor{Plum}{21}} & \textbf{\$4.69} & 5.5k & --- & \textbf{5.5k} & \textbf{\textcolor{Plum}{29}} & \textbf{\$10.54} & 9.0k & --- & \textbf{9.0k} & \textbf{\textcolor{Plum}{27}} & \textbf{\$12.93} \\
    \midrule
    \textit{\small C3\,/\,C2} & \multicolumn{3}{c}{\textit{tok:\,0.7\,\texttimes}} & \textbf{\textcolor{Plum}{21\,\texttimes}} & \textbf{cost:\,2.1\,\texttimes} & \multicolumn{3}{c}{\textit{tok:\,1.0\,\texttimes}} & \textbf{\textcolor{Plum}{29\,\texttimes}} & \textbf{cost:\,6.7\,\texttimes} & \multicolumn{3}{c}{\textit{tok:\,1.5\,\texttimes}} & \textbf{\textcolor{Plum}{27\,\texttimes}} & \textbf{cost:\,7.5\,\texttimes} \\
    \bottomrule
  \end{tabular}
  }
\end{table}

\begin{table}[t]
  \centering
  \caption{Per-task inference cost breakdown. \textit{Calls}\,=\,mean LLM calls per decision (\mbox{C3} only; \mbox{C1}/\mbox{C2} always make exactly one call); \textbf{Cost/k}\,=\,estimated USD per 1,000 decisions. \mbox{C3} call count and cost scale with task trial length: single-query tasks (PR, HT) require 11--17 calls per decision, while multi-trial tasks (BART, RB, TST) require 25--39. (OpenRouter rates, May~2026; see Table~\ref{tab:token-cost} for pricing details.)}
  \label{tab:token-cost-per-task}
  \resizebox{\linewidth}{!}{
  \begin{tabular}{l |rrrrl |rrrrl |rrrrl}
    \toprule
    & \multicolumn{5}{c}{\textbf{Qwen3-8B}} & \multicolumn{5}{c}{\textbf{Qwen3-14B}} & \multicolumn{5}{c}{\textbf{Qwen3-32B}} \\
    \cmidrule(lr){2-6} \cmidrule(lr){7-11} \cmidrule(lr){12-16}
    Task & C1\,\$/k & C2\,\$/k & \textcolor{Plum}{Calls} & C3\,\$/k & C3/C2 & C1\,\$/k & C2\,\$/k & \textcolor{Plum}{Calls} & C3\,\$/k & C3/C2 & C1\,\$/k & C2\,\$/k & \textcolor{Plum}{Calls} & C3\,\$/k & C3/C2 \\
    \midrule
    \rowcolor{black!3}Prob. Reasoning & \$0.012 & \$0.664 & \textbf{\textcolor{Plum}{11}} & \textbf{\$2.28} & \textbf{3.4\,\texttimes} & \$0.024 & \$0.338 & \textbf{\textcolor{Plum}{12}} & \textbf{\$3.52} & \textbf{10.4\,\texttimes} & \$0.019 & \$0.404 & \textbf{\textcolor{Plum}{17}} & \textbf{\$5.80} & \textbf{14.3\,\texttimes} \\
    \rowcolor{black!6}Temp. Discounting & <\$0.01 & \$0.567 & \textbf{\textcolor{Plum}{20}} & \textbf{\$4.84} & \textbf{8.5\,\texttimes} & \$0.010 & \$0.441 & \textbf{\textcolor{Plum}{22}} & \textbf{\$7.99} & \textbf{18.1\,\texttimes} & <\$0.01 & \$1.08 & \textbf{\textcolor{Plum}{38}} & \textbf{\$15.67} & \textbf{14.6\,\texttimes} \\
    \rowcolor{black!3}Horizon Task & <\$0.01 & \$0.503 & \textbf{\textcolor{Plum}{11}} & \textbf{\$1.84} & \textbf{3.7\,\texttimes} & \$0.019 & \$0.266 & \textbf{\textcolor{Plum}{21}} & \textbf{\$6.03} & \textbf{22.6\,\texttimes} & \$0.015 & \$0.400 & \textbf{\textcolor{Plum}{16}} & \textbf{\$5.26} & \textbf{13.2\,\texttimes} \\
    \rowcolor{black!6}Instr. Learning & \$0.042 & \$7.64 & \textbf{\textcolor{Plum}{18}} & \textbf{\$4.38} & \textbf{0.6\,\texttimes} & \$0.085 & \$6.67 & \textbf{\textcolor{Plum}{23}} & \textbf{\$9.20} & \textbf{1.4\,\texttimes} & \$0.068 & \$4.25 & \textbf{\textcolor{Plum}{17}} & \textbf{\$6.46} & \textbf{1.5\,\texttimes} \\
    \rowcolor{black!3}Restless Bandit & \$0.065 & \$3.92 & \textbf{\textcolor{Plum}{23}} & \textbf{\$7.33} & \textbf{1.9\,\texttimes} & \$0.129 & \$1.61 & \textbf{\textcolor{Plum}{32}} & \textbf{\$18.06} & \textbf{11.2\,\texttimes} & \$0.104 & \$3.12 & \textbf{\textcolor{Plum}{39}} & \textbf{\$24.26} & \textbf{7.8\,\texttimes} \\
    \rowcolor{black!6}BART & \$0.048 & \$1.67 & \textbf{\textcolor{Plum}{26}} & \textbf{\$7.48} & \textbf{4.5\,\texttimes} & \$0.093 & \$1.17 & \textbf{\textcolor{Plum}{34}} & \textbf{\$16.06} & \textbf{13.7\,\texttimes} & \$0.070 & \$1.71 & \textbf{\textcolor{Plum}{36}} & \textbf{\$18.95} & \textbf{11.1\,\texttimes} \\
    \rowcolor{black!3}Two-Step Task & \$0.021 & \$1.01 & \textbf{\textcolor{Plum}{20}} & \textbf{\$4.64} & \textbf{4.6\,\texttimes} & \$0.041 & \$0.460 & \textbf{\textcolor{Plum}{34}} & \textbf{\$12.94} & \textbf{28.1\,\texttimes} & \$0.033 & \$1.12 & \textbf{\textcolor{Plum}{33}} & \textbf{\$14.10} & \textbf{12.6\,\texttimes} \\
    \midrule
    \textit{Mean} & \$0.029 & \$2.28 & \textbf{\textcolor{Plum}{18}} & \textbf{\$4.69} & \textbf{2.1\,\texttimes} & \$0.057 & \$1.57 & \textbf{\textcolor{Plum}{25}} & \textbf{\$10.54} & \textbf{6.7\,\texttimes} & \$0.045 & \$1.73 & \textbf{\textcolor{Plum}{28}} & \textbf{\$12.93} & \textbf{7.5\,\texttimes} \\
    \bottomrule
  \end{tabular}
  }
\end{table}

\section{Ablation Study}
\label{app:ablations}

\begin{figure*}[t]
  \centering
  \includegraphics[width=\textwidth]{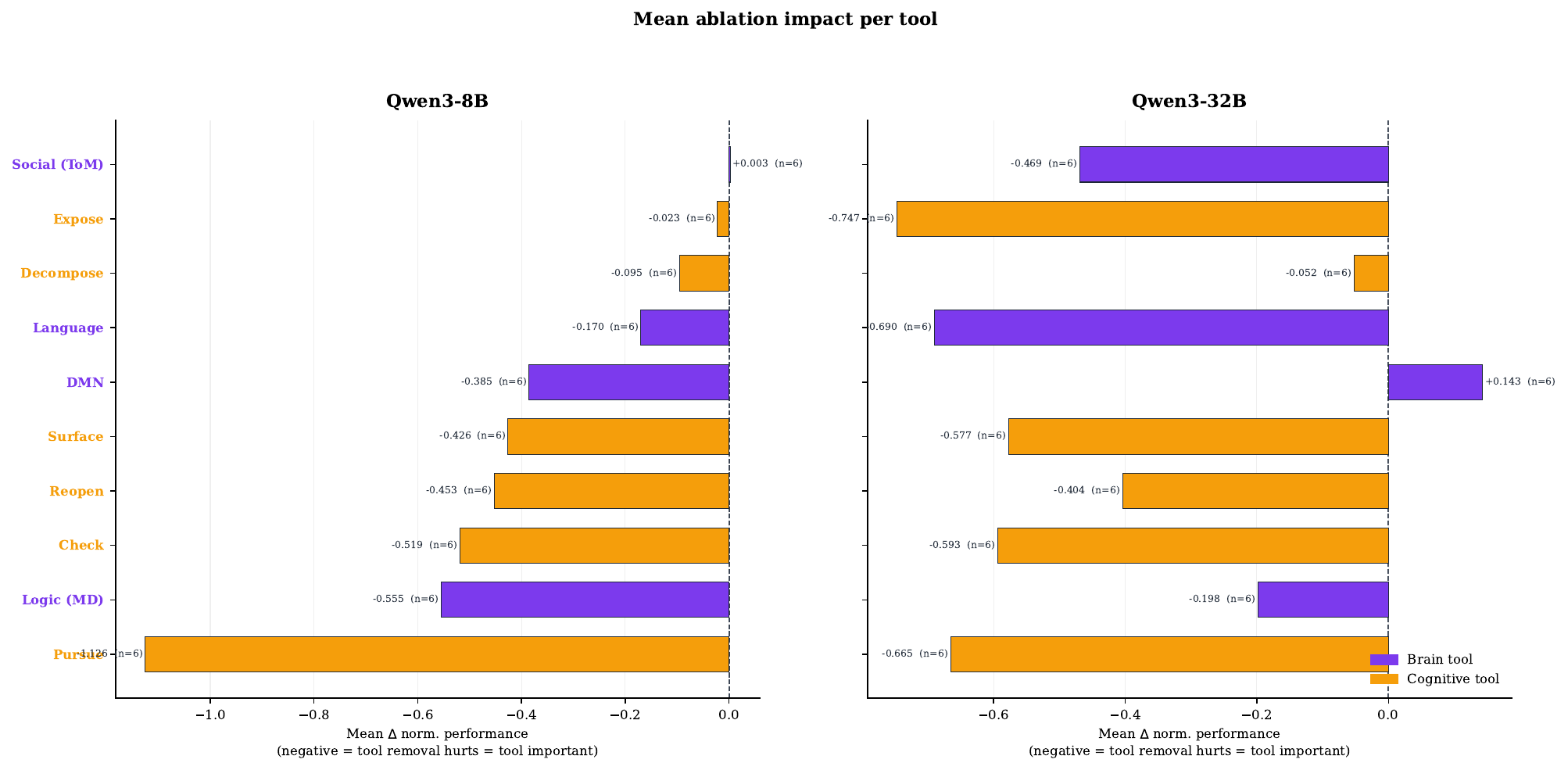}
  \caption{\textbf{Aggregate tool importance --- leave-one-out (LOO) ablation.}
  Each bar shows the mean change in normalized performance ($\Delta$, averaged across experiments)
  when one reasoning tool is removed from the full NeuReasoner (C3).
  Negative $\Delta$ means removal hurts performance (tool is load-bearing);
  positive $\Delta$ means removal helps (tool is redundant or interfering).
  Purple bars: Neuro Lenses; amber bars: Cognitive Lenses.
  Results shown for Qwen3-8B and Qwen3-32B; partial runs and Temporal Discounting (TD) excluded.
  $n$ = number of experiments contributing to each mean.}
  \label{fig:ablation_importance}
\end{figure*}

\begin{figure*}[t]
  \centering
  \includegraphics[width=\textwidth]{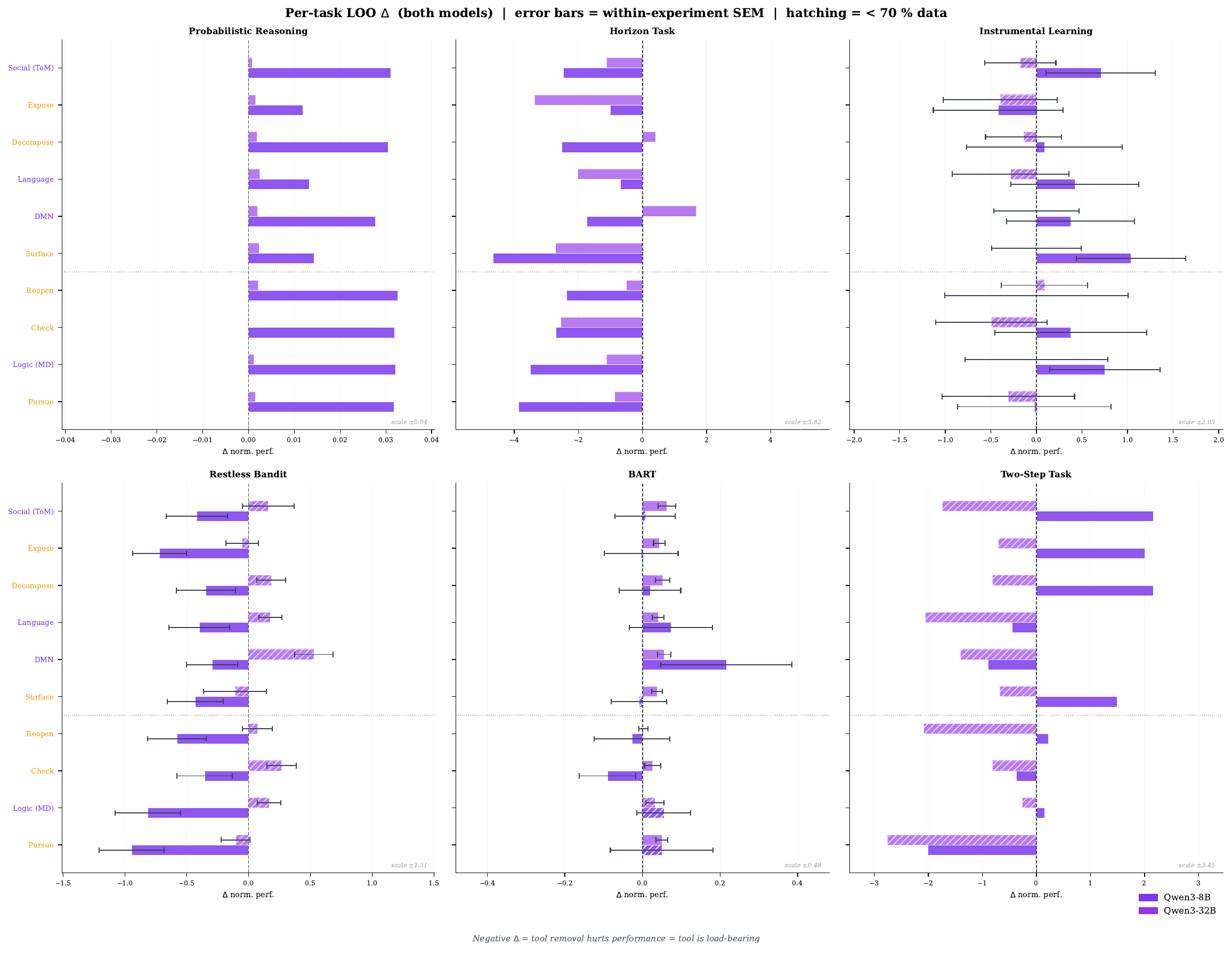}
  \caption{\textbf{Per-task LOO ablation $\Delta$ --- Qwen3-8B vs.\ Qwen3-32B.}
  Each panel shows one CogBench experiment (Temporal Discounting excluded).
  Bars compare the performance change ($\Delta$) when each tool is removed,
  side-by-side for 8B (light) and 32B (dark).
  Error bars denote within-experiment SEM; hatching marks partial runs ($<$70\% of target decisions).
  Tool ordering follows the 8B aggregate importance ranking from Figure~\ref{fig:ablation_importance}.
  Scale annotations (``$\pm X$'') indicate the per-panel axis range.}
  \label{fig:ablation_per_task}
\end{figure*}

Figures~\ref{fig:ablation_importance}--\ref{fig:ablation_per_task} summarize the leave-one-out (LOO)
ablation results; Table~\ref{tab:ablation} gives the full numerical breakdown.

\begin{figure}[t]
  \centering
  \includegraphics[width=0.60\textwidth]{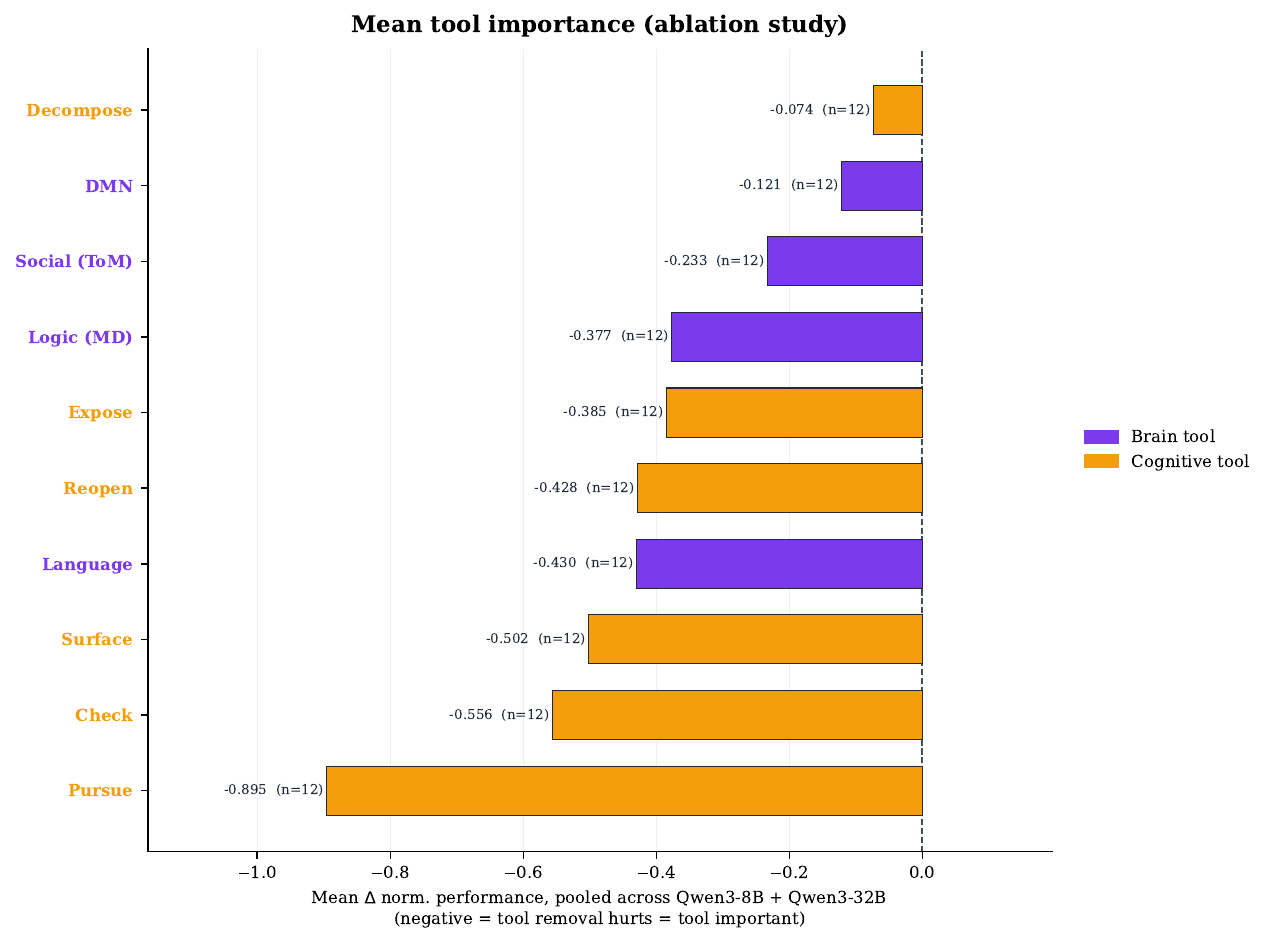}
  \caption{\textbf{Tool-ablation summary.} Mean leave-one-out
    $\Delta$\,=\,$\mathrm{norm\_perf}_{\text{ablated}} - \mathrm{norm\_perf}_{\text{full}}$,
    pooled across Qwen3-8B + Qwen3-32B and across all CogBench tasks.
    More negative values indicate a larger contribution of that tool to elicited reasoning.}
  \label{fig:ablation_summary}
\end{figure}

\begin{table}[t]
\centering
\setlength{\tabcolsep}{4pt}
\definecolor{myblue}{RGB}{37,99,235}
\caption{Leave-one-out (LOO) ablation: change in normalized performance ($\Delta$, percentage points) when one reasoning tool is removed. Rows are ranked by mean impact across models; \protect\colorbox{myred!8}{shaded} rows are the three most load-bearing tools. \textbf{Bold} = largest single-task drop per model. Red\,=\,critical; blue\,=\,redundant/interfering. B\,=\,brain lens; C\,=\,cognitive operator. HT and TST omitted ($\dagger$inflated normalization denominators make $\Delta$ values incomparable); TD omitted ($N{=}1$ run).}
\label{tab:ablation}
\resizebox{\linewidth}{!}{
\begin{tabular}{l c r r r r @{\hspace{4pt}} r @{\hspace{8pt}} r r r r @{\hspace{4pt}} r}
\toprule
& & \multicolumn{5}{c}{\textbf{Qwen3-8B}} & \multicolumn{5}{c}{\textbf{Qwen3-32B}} \\
\cmidrule(lr){3-7} \cmidrule(lr){8-12}
\textit{Tool removed} & & PR & IL & RB & BART & \textit{Avg} & PR & IL & RB & BART & \textit{Avg} \\
\midrule
\rowcolor{myred!8}\textbf{Expose} & \textcolor{gray}{\scriptsize C} & +1.2 & \cellcolor{myred!29}-42.0 & \cellcolor{myred!50}\textbf{-72.0} & \cellcolor{myred!4}-5.8 & \cellcolor{myred!21}-29.6 & +0.2 & \cellcolor{myred!50}\textbf{-79.6} & \cellcolor{myblue!12}+17.2 & +4.3 & \cellcolor{myred!10}-14.5 \\
\rowcolor{myred!8}\textbf{Pursue} & \textcolor{gray}{\scriptsize C} & +3.2 & \cellcolor{myblue!14}+19.9 & \cellcolor{myred!50}\textbf{-94.5} & +0.0 & \cellcolor{myred!12}-17.9 & +0.2 & \cellcolor{myred!43}\textbf{-61.9} & \cellcolor{myred!3}-4.3 & +5.0 & \cellcolor{myred!11}-15.3 \\
\rowcolor{myred!8}\textbf{Reopen} & \textcolor{gray}{\scriptsize C} & +3.3 & +0.0 & \cellcolor{myred!40}\textbf{-58.0} & \cellcolor{myblue!6}+9.1 & \cellcolor{myred!8}-11.4 & +0.2 & \cellcolor{myblue!12}+17.7 & +1.7 & +0.2 & +5.0 \\
Check & \textcolor{gray}{\scriptsize C} & +3.2 & \cellcolor{myblue!26}+37.6 & \cellcolor{myred!25}\textbf{-35.4} & \cellcolor{myred!6}-9.0 & \cellcolor{myred!1}-0.9 & -0.0 & \cellcolor{myred!29}\textbf{-41.3} & \cellcolor{myblue!19}+27.2 & +2.6 & \cellcolor{myred!2}-2.9 \\
Language & \textcolor{gray}{\scriptsize B} & +1.3 & \cellcolor{myblue!29}+42.0 & \cellcolor{myred!28}\textbf{-39.8} & +4.1 & +1.9 & +0.2 & \cellcolor{myred!25}\textbf{-35.4} & \cellcolor{myblue!16}+23.5 & +4.0 & \cellcolor{myred!1}-1.9 \\
Decompose & \textcolor{gray}{\scriptsize C} & +3.0 & \cellcolor{myblue!6}+8.8 & \cellcolor{myred!24}\textbf{-34.4} & \cellcolor{myred!2}-2.2 & \cellcolor{myred!4}-6.2 & +0.2 & \cellcolor{myred!4}\textbf{-5.9} & \cellcolor{myblue!24}+34.3 & \cellcolor{myblue!4}+5.2 & \cellcolor{myblue!6}+8.5 \\
Surface & \textcolor{gray}{\scriptsize C} & +1.4 & \cellcolor{myblue!50}+103.9 & \cellcolor{myred!30}\textbf{-43.0} & +1.0 & \cellcolor{myblue!11}+15.8 & +0.2 & \cellcolor{myred!12}\textbf{-17.7} & -0.2 & +4.4 & \cellcolor{myred!2}-3.3 \\
Social (ToM) & \textcolor{gray}{\scriptsize B} & +3.1 & \cellcolor{myblue!38}+55.2 & \cellcolor{myred!29}\textbf{-41.9} & \cellcolor{myred!4}-5.1 & +2.8 & +0.1 & +0.0 & \cellcolor{myblue!24}+35.2 & \cellcolor{myblue!5}+7.4 & \cellcolor{myblue!7}+10.7 \\
DMN & \textcolor{gray}{\scriptsize B} & +2.8 & \cellcolor{myblue!26}+37.6 & \cellcolor{myred!20}\textbf{-29.5} & \cellcolor{myblue!5}+6.9 & +4.4 & +0.2 & \cellcolor{myred!6}\textbf{-8.8} & \cellcolor{myblue!29}+41.2 & \cellcolor{myblue!4}+5.6 & \cellcolor{myblue!7}+9.5 \\
Logic (MD) & \textcolor{gray}{\scriptsize B} & +3.2 & \cellcolor{myblue!50}+75.1 & \cellcolor{myred!50}\textbf{-81.5} & \cellcolor{myblue!4}+5.7 & +0.6 & +0.1 & \cellcolor{myblue!49}+70.7 & \cellcolor{myblue!5}+7.5 & +1.7 & \cellcolor{myblue!14}+20.0 \\
\bottomrule
\end{tabular}
}
\end{table}

\end{document}